\documentclass[10pt,twocolumn,letterpaper]{article}
\usepackage{algorithm}
\usepackage{algorithmic}
\usepackage{cvpr}
\usepackage{times}
\usepackage{epsfig}
\usepackage{graphicx}
\usepackage{amsmath}

\usepackage{epigraph}

\usepackage[breaklinks=true,bookmarks=false]{hyperref}
\usepackage{amssymb}
\usepackage{subfig}
\cvprfinalcopy 

\def\cvprPaperID{809} 

\ifcvprfinal\fi

\begin{document}

\title{Expecting the Unexpected:\\ Training Detectors for Unusual Pedestrians with Adversarial Imposters}

\author{Shiyu Huang\\
Tsinghua University\\
Beijing, China\\
{\tt\small huangsy13@mails.tsinghua.edu.cn}
\and
Deva Ramanan\\
Carnegie Mellon University\\
Pittsburgh, USA\\
{\tt\small deva@cs.cmu.edu}
}

\maketitle
\thispagestyle{empty}

\begin{abstract}
As autonomous vehicles become an every-day reality, high-accuracy pedestrian detection is of paramount practical importance.  Pedestrian detection is a highly researched topic with mature methods,  but most datasets focus on common scenes of people engaged in typical walking poses on sidewalks. But performance is most crucial for dangerous scenarios, such as children playing in the street or people using bicycles/skateboards in unexpected ways. Such ``in-the-tail'' data is notoriously hard to observe, making both training and testing difficult. To analyze this problem, we have collected a novel annotated dataset of dangerous scenarios called the Precarious Pedestrian dataset. 

Even given a dedicated collection effort, it is relatively small by contemporary standards ($\approx 1000$ images). To allow for large-scale data-driven learning, we explore the use of synthetic data generated by a game engine. A significant challenge is selected the right ``priors'' or parameters for synthesis: we would like realistic data with poses and object configurations that mimic true Precarious Pedestrians. Inspired by Generative Adversarial Networks (GANs), we generate a massive amount of synthetic data and train a discriminative classifier to select a realistic subset, which we deem the Adversarial Imposters. We demonstrate that this simple pipeline allows one to synthesize realistic training data by making use of rendering/animation engines within a GAN framework. Interestingly, we also demonstrate that such data can be used to rank algorithms, suggesting that Adversarial Imposters can also be used for ``in-the-tail'' validation at test-time, a notoriously difficult challenge for real-world deployment. 
\end{abstract}


\section{Introduction}
\begin{figure}[t]
\begin{center}
\subfloat[]{\begin{centering}
\includegraphics[width=0.3\linewidth]{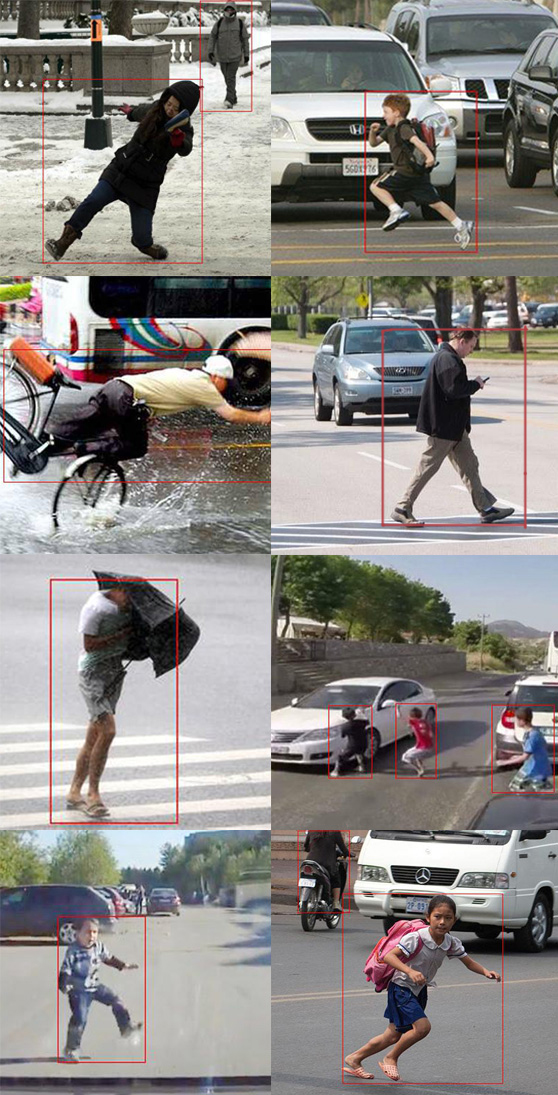}
\end{centering}
 
}\subfloat[]{\begin{centering}
\includegraphics[width=0.3\linewidth]{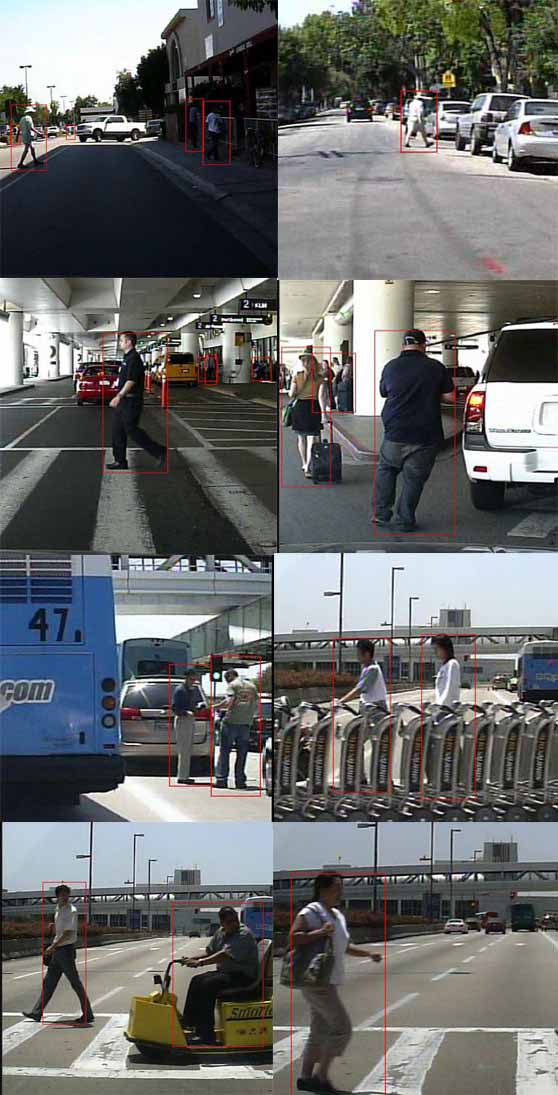}
\end{centering}
}
\subfloat[]{\begin{centering}
\includegraphics[width=0.3\linewidth]{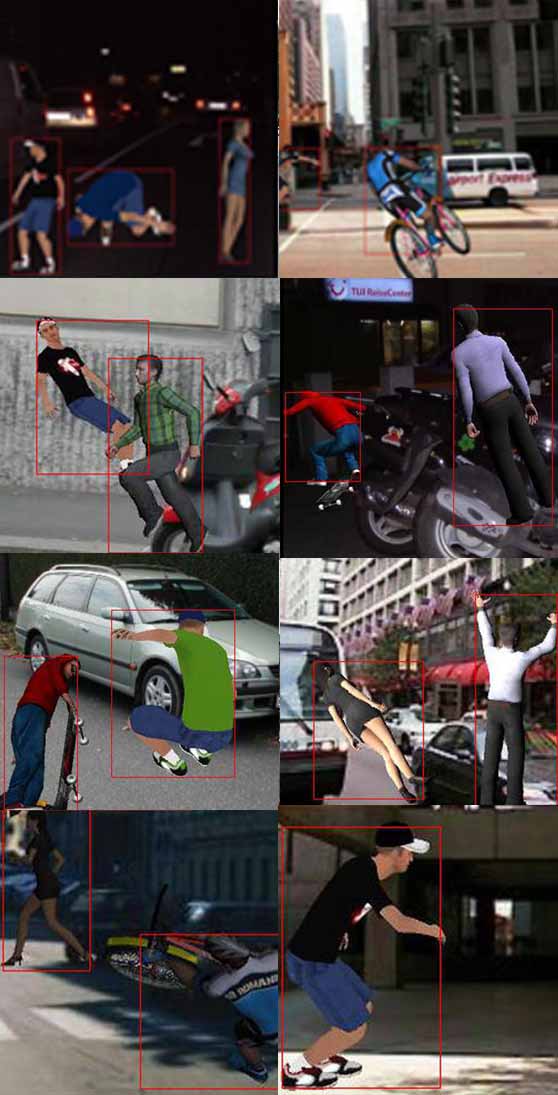}
\end{centering}
}
\end{center}
\vspace{-3ex}
\caption{(a) Examples from our novel Precarious Pedestrian Dataset of dangerous, but rare pedestrian scenes.  One important scenario is that of pedestrians on their phone (row 2, col 2), who may not be adequately aware of their surroundings. (b) Examples in Caltech Dataset tend not to capture such rare scenarios. (c) Examples from a set of Adveserial Imposters, which are synthetic images that are adversarially-trained to mimic the set of Precarious Pedstrians. We demonstrate that such images can be used to both train and evaluate robust pedestrian recognition systems targeting such dangerous scenarios.}
\label{fig:splash}
\end{figure}
\setlength{\epigraphwidth}{\linewidth}
\epigraph{\noindent \textit{There's no software designer in the world that's ever going to be smart enough to anticipate all the potential circumstances an autonomous car is going to encounter. The dog that runs out into the street, the person who runs up the street, the bicyclist, the policeman or the construction worker.}}{{\small C. Hart, Chairman of National Transport. Safety Board}}

As autonomous vehicles become an every-day reality, high-accuracy pedestrian detection is of paramount practical importance.  Pedestrian detection is a highly researched topic with mature methods,  but most datasets focus on ``everyday'' scenes of people engaged in typical walking poses on sidewalks~\cite{dollar2009pedestrian,dalal2005inria,eth_biwi_00534,enzweiler2009monocular,wojek2009multi}. However, perhaps the most important operating point for a deployable system is its behaviour in dangerous, unexpected scenarios, such as children playing in the street or people using bicycles/skateboards in unexpected ways. 

\noindent {\bf Precarious Pedestrian Dataset:} Such ``in-the-tail'' data is notoriously hard to observe, making both training and evaluation of existing systems difficult. To analyze this problem, we have collected a novel annotated dataset of dangerous scenarios called the Precarious Pedestrian Dataset. Even given a dedicated collection effort, it is relatively small by contemporary standards ($\approx 1000$ images). To explore large-scale data-driven learning, we explore the use of synthetic data generated by a game engine. Synthetic training data is an actively explored topic because it provides a potentially infinite well of annotated data for training data-hungry architectures~\cite{lerer2016learning,mayer2015large,fischer2015flownet,hattori2015learning,richter2016playing,ros2016synthia}. Particularly attractive are approaches that combine a large amount of synthetic training data with a small amount of real data (that may have been difficult to acquire and/or label).

\noindent {\bf Challenges in Synthesis:} We see two primary difficulties with the use of synthetic training data. The first is that not all data is created ``equal'': when combining synthetic data with real data, synthesizing common scenes may not be particularly useful since they will likely already appear in the training set. Hence we argue that the real power of synthetic data is generating examples ``in-the-tail'', which would otherwise have been hard to collect. The second difficulty arises in building good generative models of images, a notoriously difficult problem. Rather than building generative pixel-level models, we make use of state-of-the-art rendering/animation engines that contain an immense amount of knowledge (about physics, light transfer, etc.).  The challenge of generative synthesis then lies in constructing the right ``priors'', or scene-parameters, to render/animate.  In our case, these correspond to body poses and spatial configurations of people and other objects in the scene.

\noindent {\bf Adversarial Imposters:} We address both concerns with a novel variant of Generative Adversarial Networks (GANs)~\cite{goodfellow2014generative}, a method for synthesizing data from latent noise vectors. 
Traditional GANs learn generative feedforward models that process latent noise vectors, typically from a fixed known prior distribution. {\em Instead, we fix the feedforward model to be a rendering engine, but use an adverserial framework to learn the latent priors.} To do so, we define a rendering pipeline that takes an input a vector of scene parameters capturing object attributes and spatial layout. We use rejection sampling to construct a set of scene parameters (and their associated rendered images) that maximally confuse the discriminator. We call such examples Adversarial Imposters, and use them within a simple pipeline for adapting detectors from synthetic data to the world of real images. 

\noindent {\bf RPN+:} We use our dataset of real and imposter images to train a suite of contemporary detectors. We find surprisingly good results with a (to our knowledge) novel variant of region proposal network (RPN)~\cite{zhang2016faster} tuned for particular objects (precarious people) rather than a general class of objectness detections. Instead of classifying a sparse set of proposed windows (as nearly all contemporary object detection systems based on RCNN do~\cite{ren2015faster}), this network returns a dense heatmap of pedestrian detections, along with regressed bounding box location for each pixel location in the heatmap. We call this detector RPN+. Our experiments show that our RPN+, trained on real+imposter data, outperforms other detectors trained only on real data.

\noindent {\bf Validation:} Interestingly, we also demonstrate that our Adverserial Imposter Dataset can be used to rank algorithms, suggesting that our pipeline can also be used for ``in-the-tail'' validation at test-time, a notoriously difficult challenge for real-world deployment. 

\noindent {\bf Contributions:} The contribution of our work is as follows: (1) a novel dataset of pedestrians in dangerous situations (Precarious Pedestrians) (2) a general architecture for creating realistic synthetic data ``in-the-tail'', for which limited real data can be collected and (3) demonstration of our overall pipeline for the task of pedestrian detection using a novel detector. Our datasets and code can be found here: https://github.com/huangshiyu13/RPNplus.


\begin{figure}[t]
\begin{center}
\subfloat[]{\begin{centering}
\includegraphics[width=0.523\linewidth]{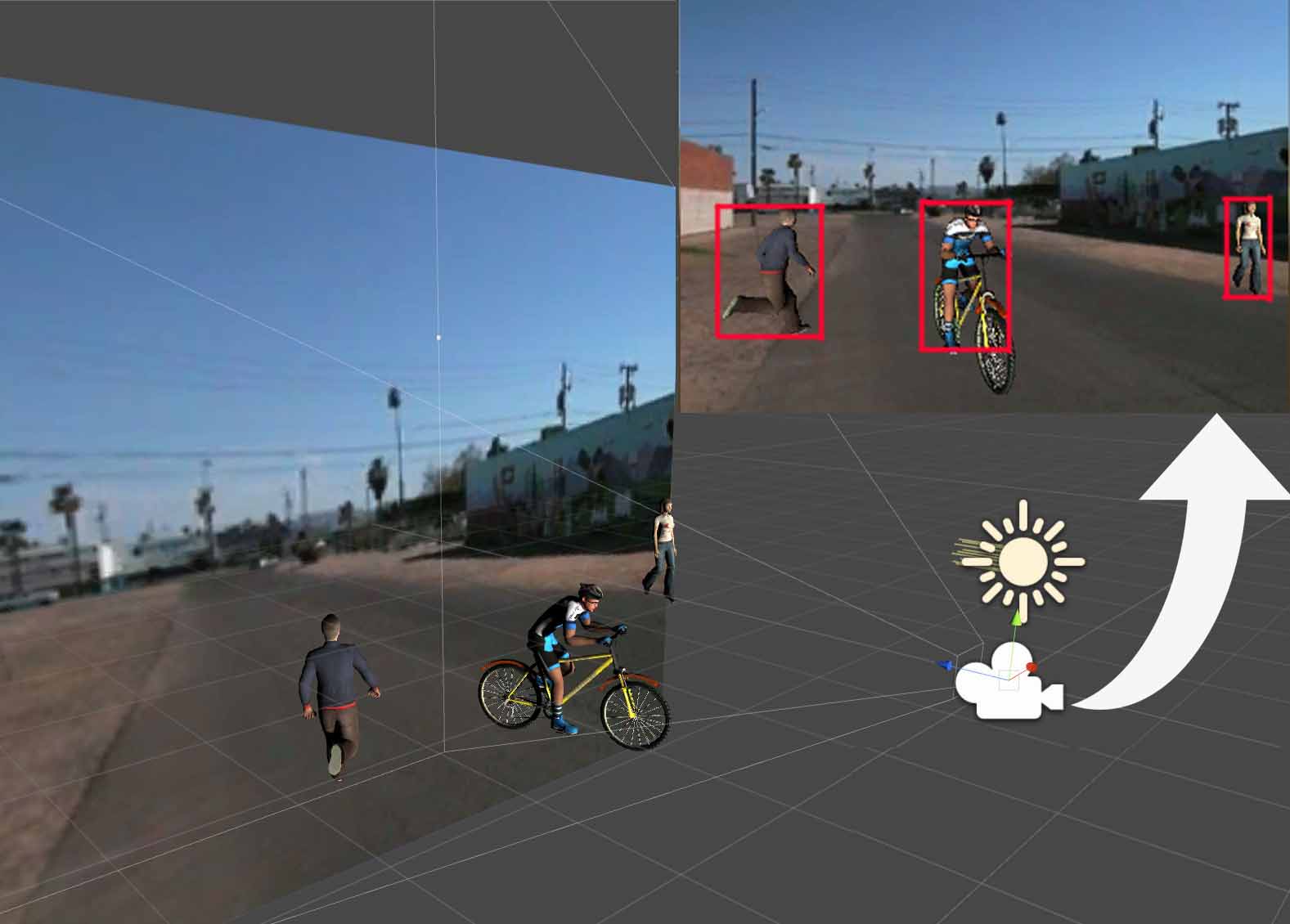}
\end{centering}
 
}\subfloat[]{\begin{centering}
\includegraphics[width=0.4\linewidth]{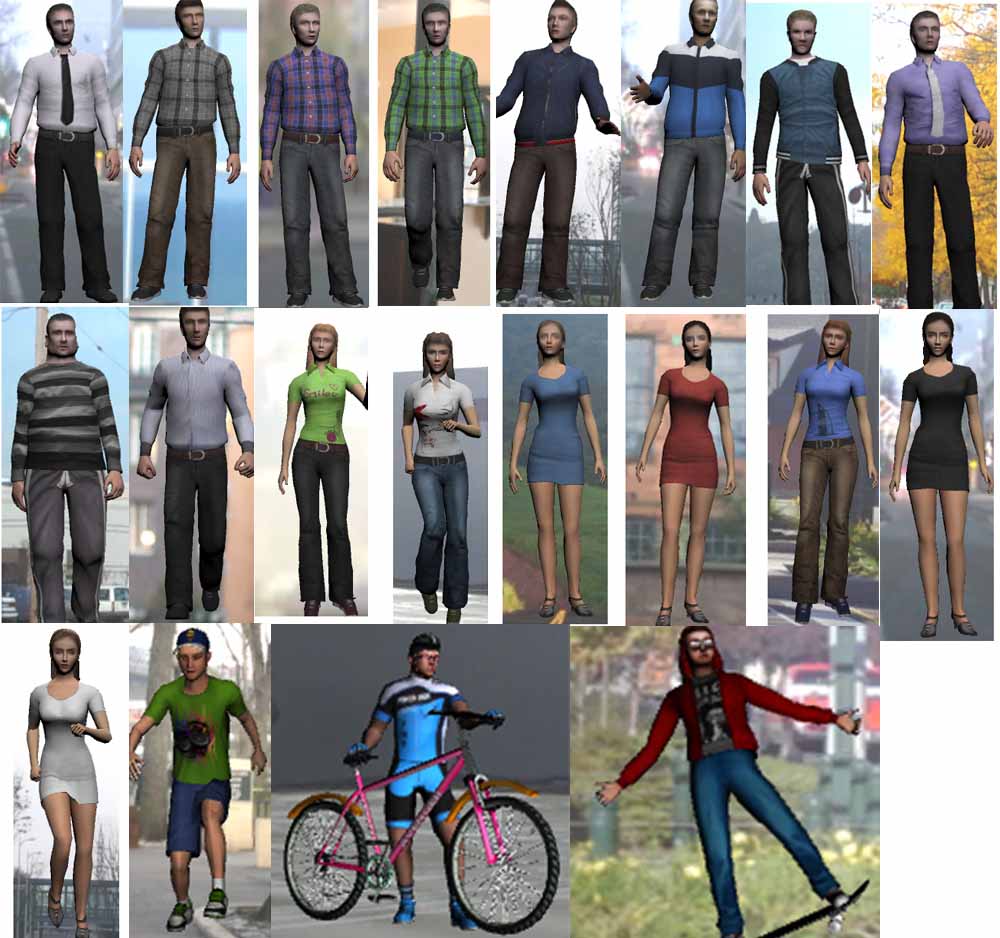}
\end{centering}
}
\end{center}
\vspace{-3ex}
   \caption{(a) Scene that's built for generating synthetic images. (b) 3D models that we use in this project.}
\label{fig:generateSynData}
\end{figure}
\section{Related work}

\noindent
{\bf Synthetic Data:} Synthetic datasets have been used to train and evaluate the performance of computer vision algorithms. Some forms of ground truth are hard to obtain from hand-labelling, such as optical flow, but easy to synthesize via simulation~\cite{fischer2015flownet}. Adam~\etal~\cite{lerer2016learning} used a 3D game engine to generate synthetic data and learned an intuitive physics model to predict falling towers of blocks. Mayer~\etal~\cite{mayer2015large} released a benchmark suite of various tasks using synthetic data, including  disparity and optical flow. Richter \etal~\cite{richter2016playing} used synthetic data to improve image segmentation performance, but notably do not control the scene as to explore targeted arrangements of objects. 
German Ros \etal~\cite{ros2016synthia} used Unity Development Platform to generate a synthetic urban scene dataset.\\
\noindent
{\bf 3D Models for Detection:} A notable application of 3D computer graphics model in vision has been the modeling of the human body shapes~\cite{grauman2003inferring,broggi2005model,agarwal2006local,marin2010learning,potamias2008nearest,athitsos2010database,romero2010hands}. Moreover, 3D simulation can also been used for car detection~\cite{pepik2012teaching,movshovitz20143d,hejrati2014analysis} and scene understanding~\cite{satkin2012data,lai2014unsupervised}. Marin~\etal~\cite{marin2010learning} used a game engine to generate synthetic training data. Pishchulin~\etal~\cite{pishchulin2011learning} used 8 HD cameras to scan human body and built real 3D human models. Then they used synthetic data and some labelled real data to train pedestrian detectors. Hattori \etal~\cite{hattori2015learning} used 3D modelling software to build a special scene and randomly put 3D models on a special background for pedestrian detection. Most of these works use synthetic data ``as-is'', while we analyze statistical differences between synthetic and real data, describing a pipeline for reconciling such differences through adversarial domain adaption. \\
\noindent
{\bf Domain Adaptation:} Domain Adaptation is a standard strategy to deal with data across different domains, such as synthetic versus real. Large synthetic datasets can be used to bootstrap detectors and then adapted to real data by moving to the target domain distribution. Sun and Saenko~\cite{sun2014virtual} used 3D models to train detectors for real objects. 
Such work typically used shallow detectors defined on fixed feature sets, while we focus on gradient-based adaption of ``deep'' detection networks (such as RCNN).  From this perspective, our work is inspired by approaches for deep domain adaptation~\cite{ganin2014unsupervised,ganin2016domain,liu2016coupled,long2015learning}. Such work typically assumes that one has access to large amounts of unlabeled data from the target domain. In our case, assembling a large target dataset of unlabelled examples (of real Precarious Pedestrians) is {\em itself} challenging, necessitating the need for alternative approaches that make stronger use of the source dataset. 

\noindent
{\bf Generative Adversarial Nets:} GANs~\cite{goodfellow2014generative}  are deep networks that can generate synthetic images from latent noise vectors. They do so by adversarially-training a neural network to discriminate between real versus synthetic images. Recent works have shown impressive performance in generation of synthetic images~\cite{mirza2014conditional,denton2015deep,radford2015unsupervised,salimans2016improved,chen2016infogan}. However, it appears challenging to synthesize high-resolution images with semantically-valid content. We circumvent these limitations with a rendering-based adversarial approach to image synthesis. 


\begin{figure}[t]
\begin{center}
\subfloat[Precarious Dataset]{\begin{centering}
\includegraphics[width=0.5\linewidth]{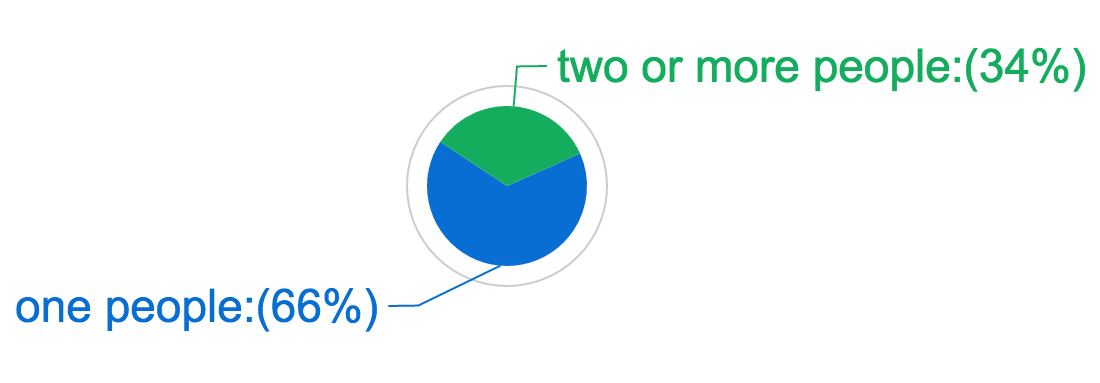}
\end{centering}
}\subfloat[Caltech Dataset]{\begin{centering}
\includegraphics[width=0.5\linewidth]{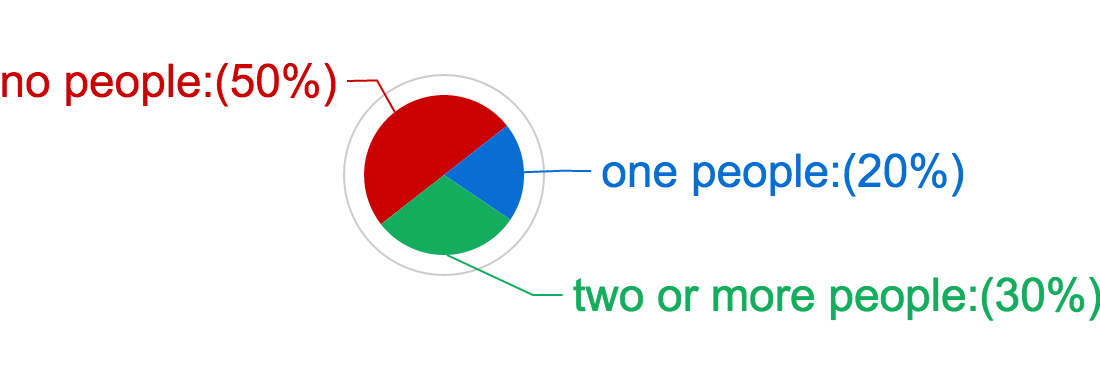}
\end{centering}
}

\subfloat[Precarious Dataset]{\begin{centering}
\includegraphics[width=0.5\linewidth]{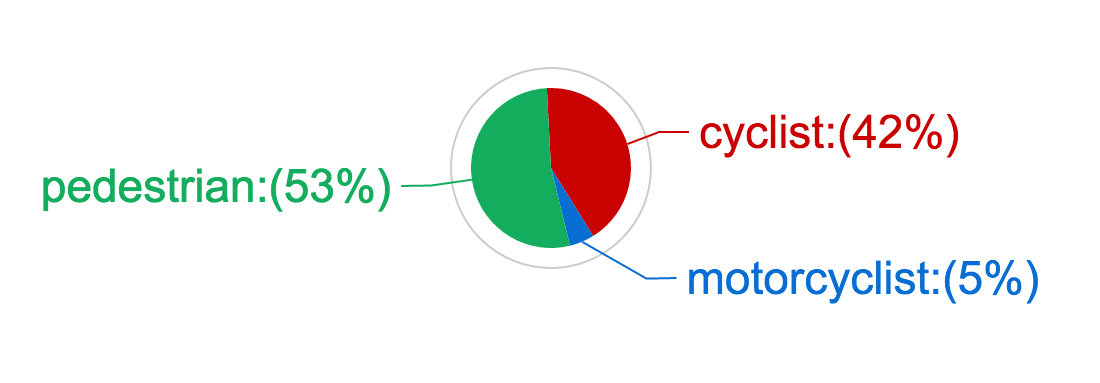}
\end{centering}
}\subfloat[Caltech Dataset]{\begin{centering}
\includegraphics[width=0.5\linewidth]{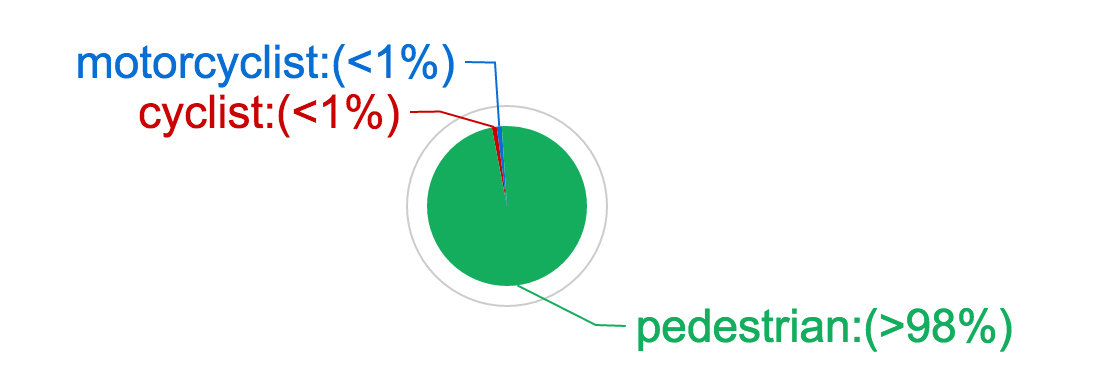}
\end{centering}
}
\end{center}
\vspace{-3ex}
\caption{(a) and (b) show the percentage of the number of people per image in both datasets. (c) and (d) show the percentage of the different types of people in both datasets. Precarious Dataset contains more cyclists and motorcyclists than Caltech Dataset.}
\label{fig:dan}
\end{figure}

\section{Datasets}

\subsection{Precarious Pedestrian Dataset}
We begin by describing our Precarious Pedestrian Dataset. We perform a dedicated search for targeted keywords (such as ``pedestrian fall", ``traffic violation" and ``dangerous bike rider") on Google Images, Baidu Images, and some selected images from MPII Dataset~\cite{andriluka14cvpr}, producing a total of 951 reasonable images. We then label bounding boxes for each image manually. Precarious Pedestrians contains various kinds scenes, such as children running on the road, people tripping, motorcyclists performing dangerous movements, people interacting with objects (such as bicycles or umbrellas). One important dangerous but increasingly common scenario consists of people watching their phones or texting while crossing the street, which is potentially dangerous as the person may not be aware of their surroundings (Figure \ref{fig:splash}). To quantify the (dis)similarity of Precarious Pedestrians to standard pedestrian benchmarks such as Caltech~\cite{Dollar2012PAMI}, we tabulate the percentage of images with more than one people, as well as the number of irregular ``pedestrians'' such as bicyclists or motorcyclists. Compared to Caltech, Precarious Pedestrians contains images with many more overall people as well as many more cyclists and motorbikes (Figure~\ref{fig:dan}). We split the Precarious dataset equally for training and testing.


\subsection{Synthetic Dataset}
\label{sec:syn}
To help both train and evaluate algorithms for detecting precarious pedestrians, we make use of a synthetic data. In this section, we describe our rendering pipeline for generating synthetic data. 
We use the Unity 3D game engine as our basic platform for simulation and rendering, due to the large availability of both commercial and user-generated {\em assets}, in the form of 3D models and character animations. \\

\indent Figure \ref{fig:generateSynData} shows the commercial 3D human models that we use for data generation, consisting of 20 models spanning different women, men, cyclists and skateboarder avatars. Because these are designed for game engine play, each 3D model is associated with characteristic animations such as jumping, talking, running, cheering and applauding. We animate these models in a 3D scene with a 2D {\em billboard} to capture the scene background~\cite{eberly20063d}, as shown in Figure~\ref{fig:generateSynData}.   Billboards are randomly sampled from the 1726 background images from INRIA dataset~\cite{dalal2005inria} and a custom set of outdoor scenes downloaded from Internet.  Our approach can generate a diverse set of background scenes, unlike approaches that are limited to a single virtual urban city~\cite{marin2010learning}.\\\\
\begin{table}
\begin{center}

\begin{tabular}{|c|c|}
\hline
 & Range \\
\hline\hline
Number of 3D models        & $[4,8]$ \\
Index of background images & $[0,1726)$ \\
Index of 3D models         & $[0,20)$ \\
Position of 3D models      & Within the field of vision \\
Index of Animations        & $[0, maxnumber)$ \\
Time of animation          & $[0,1]$\\
Model's angle on the x axis& $[-90^{\circ},90^{\circ}]$\\
Model's angle on the y axis& $[-180^{\circ},180^{\circ}]$\\
Model's angle on the z axis& $[-90^{\circ},90^{\circ}]$\\
Light intensity            & $[0.5,2]$\\
Light's angle on the x axis& $[-45^{\circ},45^{\circ}]$\\
Light's angle on the y axis& $[-45^{\circ},45^{\circ}]$\\
\hline
\end{tabular}

\end{center}
\vspace{-3ex}
\caption{Constraints of parameters for synthesizing images. The index and time(normalized) of animations will jointly decide the gestures of 3D models .}
\label{table:constraints}
\end{table}

\noindent
{\bf Scene parameters:} To build a large library of synthetic images that will potentially be used for training and evaluation, we first define a set of parameters and parameter ranges.
We index the set of background images, the set of 3D models, and the animation frame number for each model.
In brief, the scene parameters include directional light intensity and direction (capturing sunlight), the background image index, the number of 3D models, and for each model, an index specifying the avatar ID and animation frame, as well as a root position and orientation (rotation in the ground plane). We assume a fixed camera viewpoint.
Note that the root position affects both the location and scale of the 3D model in the rendered image.
All these parameters can be summarized  as a variable-length vector $\mathbf{z} \in \mathcal{Z}$, where each vector corresponds to a particular scene instantiation.

\noindent {\bf Synthesis:} Our generator $G({\bf z})$, or rendering engine, synthesizes an image corresponding to ${\bf z}$. Importantly, we can also synthesize labels $L({\bf z})$ for each rendered image, specifying object type, 3D location, pixel segmentation masks, \etc. In practice, we make use of only 2D object bounding boxes.
Table \ref{table:constraints} shows the viable ranges of each parameter. In addition, we found the following heuristic to simulate reasonable object layouts: we enforce the maximum overlap between any two 3D models to be 20\% (to avoid congestion) and the projected location of the 3D models should lie within the camera's field-of-view. These conditions are straightforward to verify for a given vector  $\mathbf{z}$ without rendering any pixels, and so can be efficiently enforced though rejection sampling (i.e., generate a random vector and only render those that pass these conditions). Unlike Hironori~\etal\cite{hattori2015learning}, who generate training data by manually tuning $\mathbf{z}$ to match specific scenes, our approach is not scene specific and does not require any manual intervention.

\noindent {\bf Pre-processing:} Synthesized images and Precarious Pedestrian images may be of different sizes. We isotropically scale each image to a resolution of  960$\times$720, zero-padding as necessary. Our experiments also make use of the Caltech Pedestrian benchmark, to which we apply the same pre-processing.

\begin{figure}[t]
\begin{center}
\subfloat[]{\begin{centering}
\includegraphics[width=0.5\linewidth]{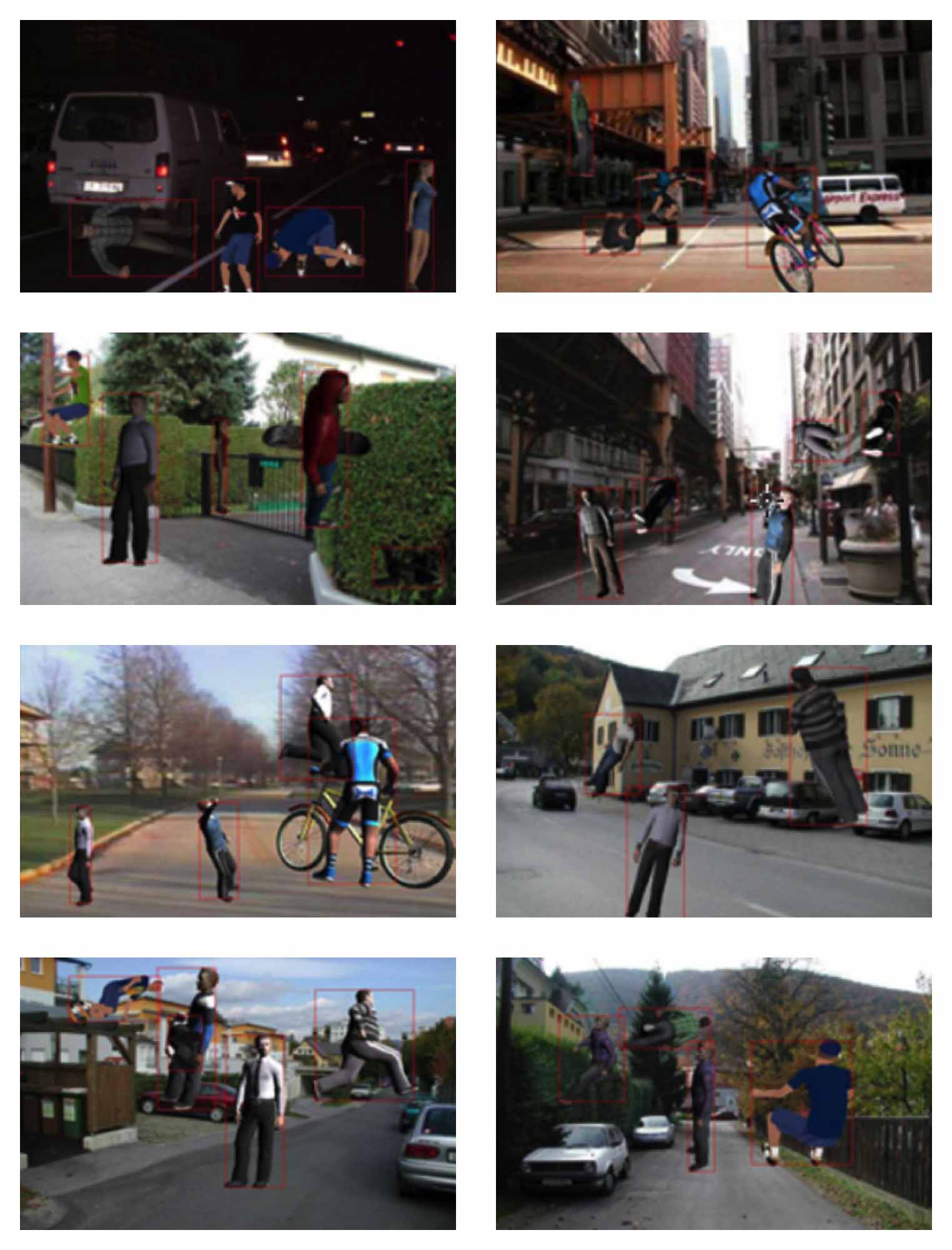}
\end{centering}
}\subfloat[]{\begin{centering}
\includegraphics[width=0.5\linewidth]{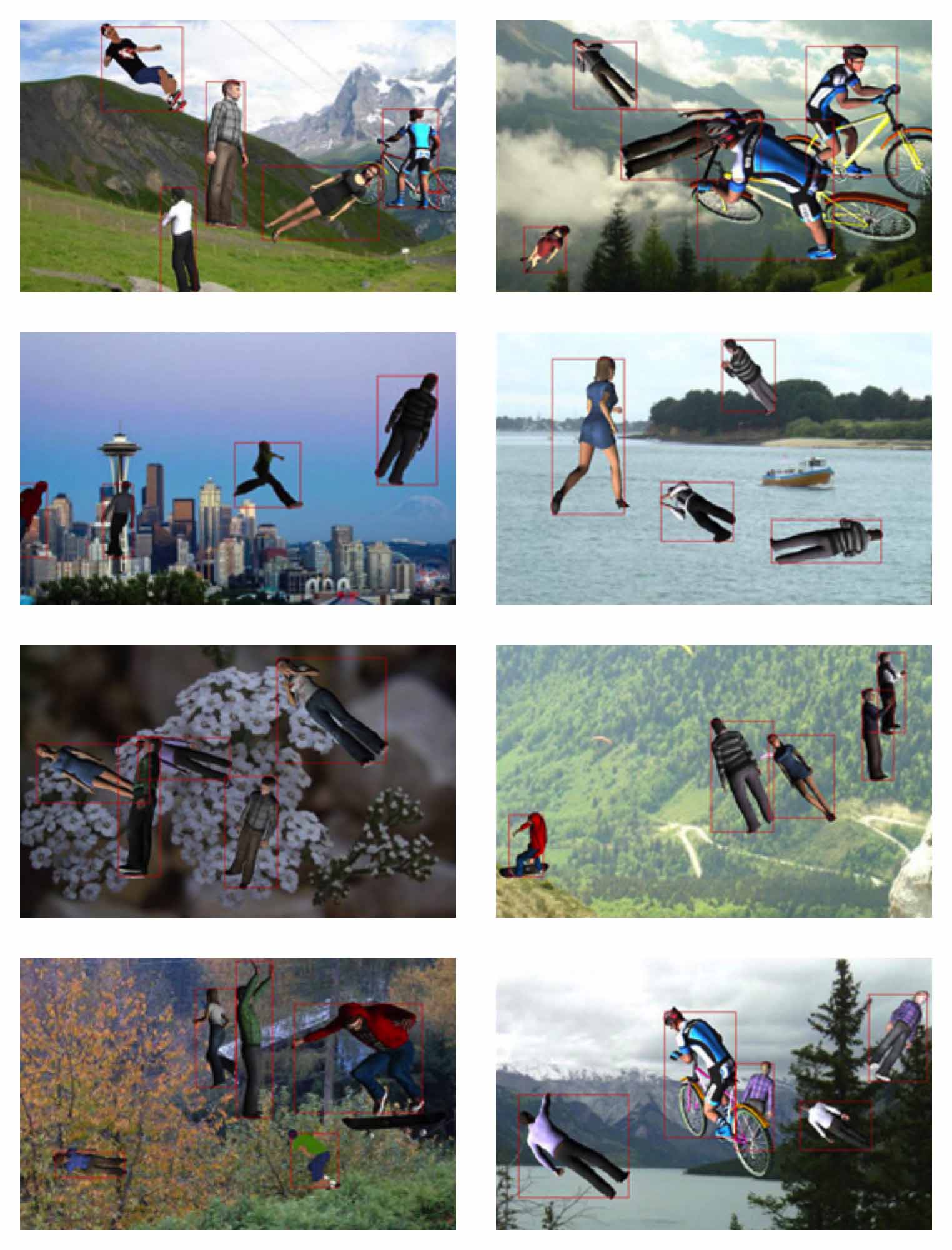}
\end{centering}
}
\end{center}
\vspace{-3ex}
   \caption{(a) Imposter images that are chosen by selector. (b) Synthetic images that are not in Imposter Dataset.}
\label{fig:impostersExamples}
\end{figure}


\section{Proposed Method}
\noindent {\bf Domain adaption:} In this section, we introduce a novel framework for adversarially adapting detectors from synthetic training data to real training data. 
We use $\mathbf{x} \in \mathcal{X}$ to denote an image and $\mathbf{y} \in \mathcal{Y}$ to denote its label vector (a set of bounding box labels). Let $p_s({\bf x}, {\bf y})$ to refer to the distribution of image-label pairs from the source domain (of synthetic images), and $p_t({\bf x},{\bf y})$ to refer to the target domain (of real Precarious Pedestrians). In our problem, we expect large amounts of source samples, but a limited amount of target ones. We factorize the joint into a marginal over image appearance and conditional on label given the appearance - e.g., $p_s({\bf x})p_s({\bf y}|{\bf x})$. Importantly, we discriminatively train a feedforward function $ f_s({\bf x}) = p_s({\bf y}|{\bf x})$ to match the conditional distribution. Our central question is how to transfer feedforward predictors trained from source samples $f_s({\bf x})$ to the target domain $f_t({\bf x})$.

\noindent {\bf  Fine-tuning:} The most natural approach to domain adaption may simply be to fine-tune a predictor $f_s({\bf x})$, originally trained on the source, with samples from the target $p_t({\bf x},{\bf y})$. Indeed, virtually all contemporary methods for visual recognition makes use of fine-tuned models that were pre-trained on Imagenet~\cite{russakovsky2015imagenet}. We compare to such a strategy in our experiments, but find that fine-tuning works best when source and target distributions are similar.  As we argue, while rendering engines can produce photorealistic scenes, it is difficult to specify a prior over scene parameters that mimic real (Precarious) scenes. We describe a solution that adversarially learns a prior. 

\noindent {\bf Generators:} As introduced in Sec.~\ref{sec:syn}, let ${\bf z} \in \mathcal{Z}$ be a vector of scene parameters, $G({\bf z}) \in \mathcal{X}$ be a feedforward generator function that renders a synthetic image given the scene parameters, and $L({\bf z}) \in \mathcal{Y}$ be a function that generates labels from the scene parameters. We can then {\em reparameterize} the distribution over synthetic images as a distribution over scene parameters $p_z({\bf z})$. We now describe a procedure for learning a prior $p_z({\bf z})$ that allows for easier transfer. Specifically, we learn a prior that {\em fools an adversary }that is trying to distinguish samples from the source and target.

\noindent {\bf Adversarial generators:}
To describe our approach, we first recall a traditional generative adverserial network (GAN):
\begin{align}
&\min_G \max_D V(D,G) = \qquad \text{[Gen. Adversarial Net]}\\
&\mathbb{E}_{\mathbf{x}\thicksim p_t(\mathbf{x})} [\log D({\bf x})] + \mathbb{E}_{\mathbf{z}\thicksim p_z(\mathbf{z})} [\log (1 - D(G({\bf z})))] \nonumber
\end{align}
where the minmax optimization jointly tries to estimate a discriminator $D$ that can distinguish real versus synthesized data examples, and the generator $G$ tries to synthesize realistic examples that fool the discriminator.  
Typically, the discriminator $D({\bf x})$ is trained to output the probability that ${\bf x}$ is real (e.g., a real Precarious Pedestrian), while  $p_z({\bf z})$ is fixed to be a zero-mean, unit-variance Gaussian. This optimization can be performed with stochastic gradient updates, that converge (in the limit) to a fixed point of the minimax problem. We refer the reader to the excellent introduction in~\cite{goodfellow2014generative}. Importantly, the generator must encode complex constraints about the manifold of natural images, that capture amongst other knowledge the physical properties of light transport and material appearance. 

\noindent {\bf Adversarial priors:} We note that {\em rendering engines} can be viewed as generators that already contain much of this knowledge, and so we fix $G$ to be a production-quality rendering platform (Unity 3D). Instead,  we learn the {\em prior} over parameter vectors in a adversarial manner:
\begin{align}
  &\min_I \max_D V(D,I) = \qquad  \text{[Adversarial Priors]}  \label{eq:adprior}\\
&\mathbb{E}_{\mathbf{x}\thicksim p_t(\mathbf{x})} [\log D({\bf x})] + \mathbb{E}_{\mathbf{z}\thicksim p_I(\mathbf{z})} [\log (1 - D(G({\bf z})))] \nonumber
\end{align}
If the generator $G$ is differentiable with respect to $z$, it is possible to use backprop to compute gradient updates for simple prior distributions $p_I({\bf z})$, such as Gaussians ~\cite{kingma2013auto,rezende2014stochastic}. This implies that the above formulation of adversarial priors is amenable to gradient-based learning.

\noindent {\bf Imposter search:}  We see two difficulties with directly applying \eqref{eq:adprior} to our problem: (1) It seems unlikely that the optimal prior for precarious scene parameters will be a simple unimodal distribution with a single mean parameter vector (and associated covariance matrix). (2) Rendering, while readily expressed as a feed-forward function, is not naturally differentiable at object boundaries (where small changes in parameters can generate large changes in the rendered image). While approximate differentiable renderers do exist~\cite{loper2014opendr}, we wish to make use of highly-optimized commercial packages for animation and image synthesis (such as Unity 3D). As such, we adopt a simple sampling-based approach that addresses both limitations:
\begin{align}
  &\min_I \max_D V(D,I) = \qquad  \text{[Imposter Selection]}\\ 
&\mathbb{E}_{\mathbf{x}\thicksim p_t(\mathbf{x})} [\log D({\bf x})] + \mathbb{E}_{\mathbf{z}\thicksim Unif(\mathcal{Z}_I)} [\log (1 - D(G({\bf z})))] \nonumber
\end{align}
\noindent where $\mathcal{Z}_I \subseteq \mathcal{Z}$. That is, we search for a subset of parameter vectors (the ``imposters'') that fool the discriminator. One could employ various sequential sampling strategies for optimizing the above; start with a random sample of parameter vectors, update the discriminator (with gradient based updates using a batch of real and synthesized data), generate additional samples close to those imposters that fool the discriminator, and repeat. We found a single iteration to work quite well. Our algorithm for synthesizing a realistic set of precarious scenes is given in Alg.~\ref{algo:selector}, and the overall approach for advesarial domain adaption is given in Alg.~\ref{algo:adapt}.
\begin{algorithm}[h!]
\caption{Imposter Selection}
\label{algo:selector}
\begin{algorithmic}
\STATE {\bfseries Input:} Set of examples from source domain $S$ and target domain $T$.\\
\STATE {\bfseries Output:} Subset of imposters $I \subseteq S$.\\
 \STATE{\bfseries 1. Train} a binary discriminator network $D({\bf x})$ that distinguishes examples ${\bf x} \in S$ from ${\bf x} \in T$.\\
 \STATE{\bfseries 2. Return} the subset of $k$ samples from $S$ that best fool the discriminator.\\
\end{algorithmic}
\end{algorithm}

\begin{algorithm}[h!]
\caption{Domain Adaption with Imposters}
\label{algo:adapt}
\begin{algorithmic}
\STATE {\bfseries Input:} Set of examples from source domain $S$ and target domain $T$.\\
\STATE {\bfseries Output:} Predictor $f({\bf x})$ for target set $T$.\\
 \STATE{\bfseries 1. Pre-Train}  a predictor $f({\bf x})$ on source set $S$.\\
 \STATE{\bfseries 2. Adapt} the predictor on $T \cup I$, where $I$ is the set of imposters found with Alg.~\ref{algo:selector}.\\
 \STATE{\bfseries 3. Fine-tune} the predictor on only target set $T$.\\
\end{algorithmic}
\end{algorithm}

Here, the set $S$ consists of synthetic image-label pairs rendered from an exhaustive set of scene parameters $\{z\}$ and the set $T$ consists of real (Precarious) image-label pairs. Without Step 2, Alg.~\ref{algo:adapt} reduces to standard fine-tuning from a source to a target domain.  Step 2 can be thought of as ``marginal distribution adaption'', since the distribution of imposter images $p_I({\bf x})$ mimics the true target distribution $p_t({\bf x})$, at least from the discriminator's perspective. 
But importantly, the discriminator $D({\bf x})$ has not made use of labels ${\bf y}$ to find imposters, and so imposter labels may not mimic the true target label distribution. Because of this, we opt to finally fine-tune $f({\bf x})$ on the target image-label pairs. Alternatively, one may explore a discriminator that directly operates on pairs of data and labels, as in~\cite{isola2016image}.

\begin{figure}[t!]
\begin{center}
   \includegraphics[width=1\linewidth]{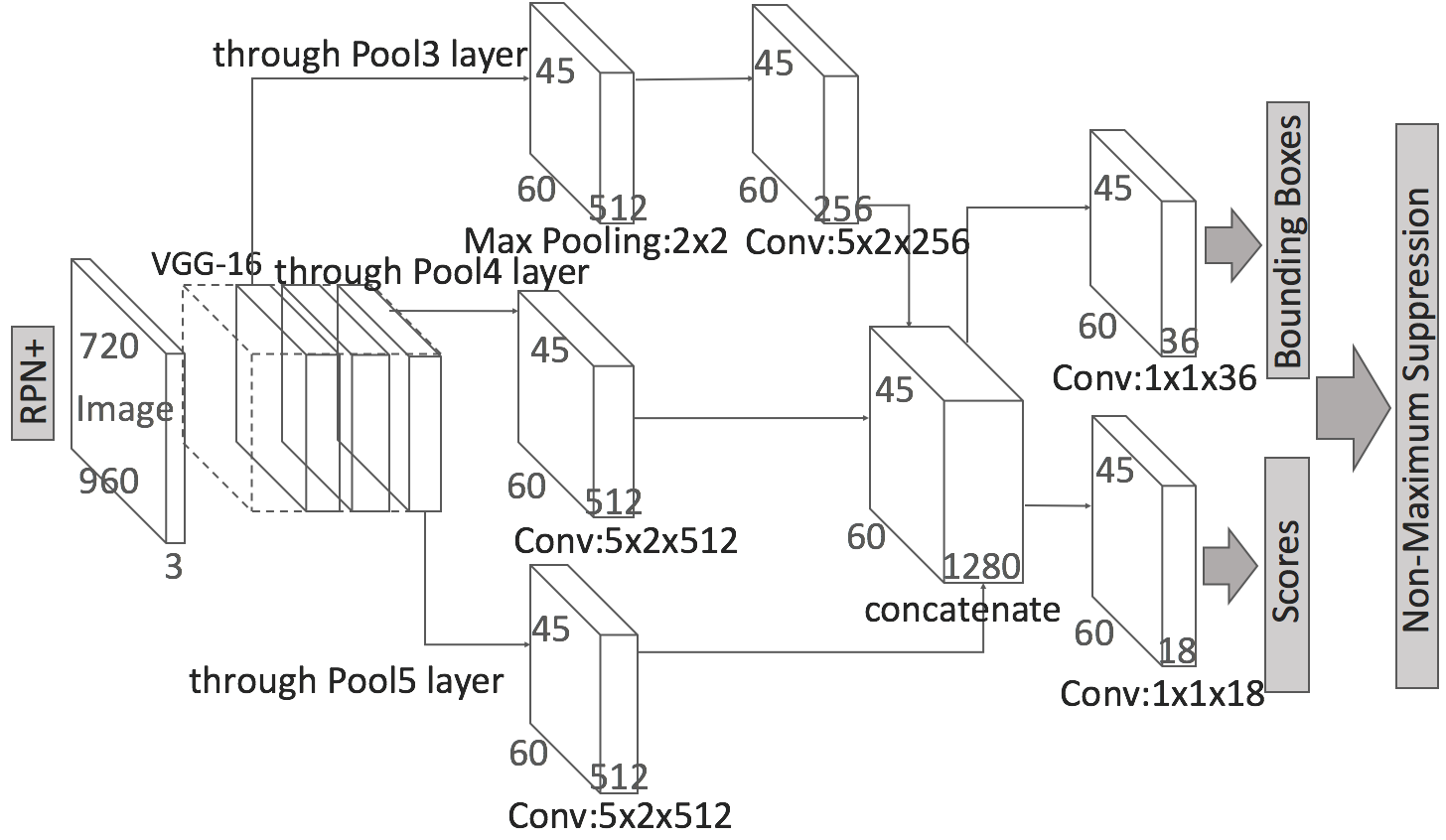}
\end{center}
\vspace{-3ex}
   \caption{Architecture of RPN+.}
\label{fig:networks}
\end{figure}

\subsection{Implementation}

\noindent {\bf Discriminator $D({\bf x})$:} Our discriminator $D$ is a VGG16 network trained to output the probability that an input images is real (with label 1) or synthetic (with label 0). 
We found that modest number of images sufficed for training: 500 images from the Precarious Pedestrian train split and 1000 random synthetic images. We downsample images to 384$\times$288 to accelerate training. After training $D$, we generate another set of 8000 synthetic images and select various subsets of size $k$ to define the imposter set (examined further in our experiments). We roughly find that 2.5\% of the synthesized images can serve as reasonable imposters.

\noindent {\bf Predictor $f({\bf x})$:} We make use of a detection system based off on a region proposal network (RPN)~\cite{ren2015faster,zhang2016faster}. Rather than training a RPN to return objectness proposals, we train it to directly return pedestrian bounding boxes.  Our network, denoted as RPN+, is illustrated in Figure \ref{fig:networks}. RPN+ is a fully convolutional network implemented with TensorFlow. We concatenate several layers on different stages in order to improve the ability of locating people in different resolutions. We use 9 anchors (reference boxes with 3 scales and aspect ratios) at each sliding position. During training, a candidate bounding box will be treated as a positive if its intersection-over-union overlap with a ground-truth box exceeds 50\%, and will be a negative for overlaps less than 20\%.
To accelerate training time, we initialize with a pre-trained VGG-16 model where the first two convolutional layers are frozen. 


\section{Experiments}
\subsection{Evaluation}
We follow the evaluation protocol of the Caltech pedestrian dataset~\cite{Dollar2012PAMI}, which use ROC curves for 2D bounding box detection at 50\% and 70\%  overlap thresholds. 

\noindent {\bf Testsets:} We use three different datasets for evaluation:  our novel {\bf Precarious Pedestrian} testset of real images,  our novel  {\bf Adverserial Imposter Testset}, and for diagnostics, a standard pedestrian benchmark dataset ({\bf Caltech}).

\noindent {\bf Baselines:} We compare our approach with the following baselines:\\
\noindent
{\bf ACF:} An aggregate channel features detector~\cite{dollar2014fast} . \\
\noindent
{\bf LDCF:} A LDCF detector~\cite{nam2014local}.\\ 
{\bf HOG+Cascade:} A cascade of boosted classifiers working with HOG features~\cite{zhu2006fast}.\\
\noindent
{\bf HARR+Cascade:}  A cascade of boosted classifiers working with haar-like features\cite{viola2001rapid,lienhart2002extended}.\\
\noindent
{\bf RPN/BF:} A RPN detection model trained with boosted forest~\cite{zhang2016faster}, which appears to be the state-of-the-art pedestrian detection system at the time of publication.\\
\noindent
\noindent

\begin{figure}[t!]
\begin{center}
\subfloat[Precarious Dataset]{\begin{centering}
\includegraphics[width=0.5\linewidth]{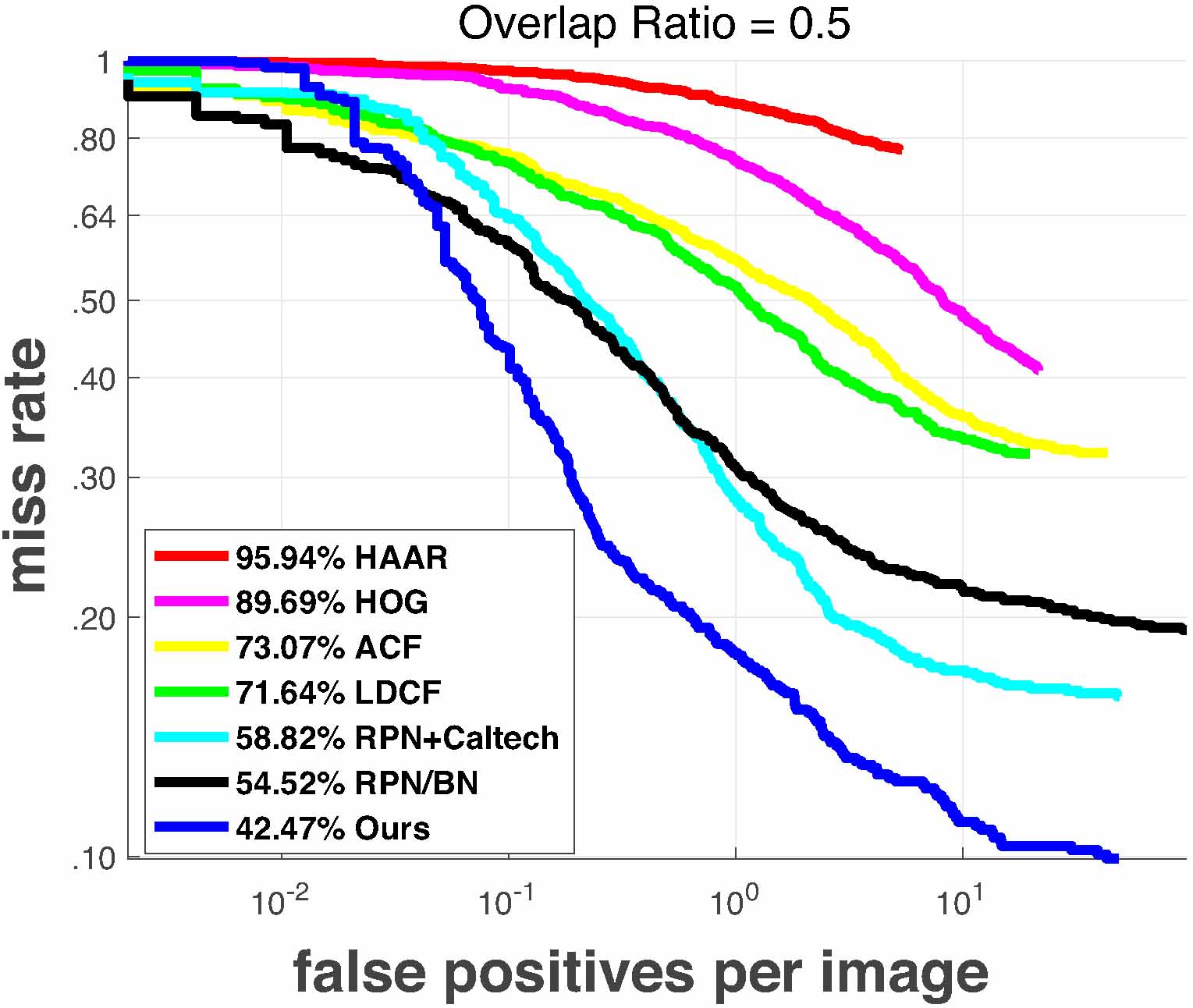}
\end{centering}
}
\subfloat[Precarious Dataset]{\begin{centering}
\includegraphics[width=0.5\linewidth]{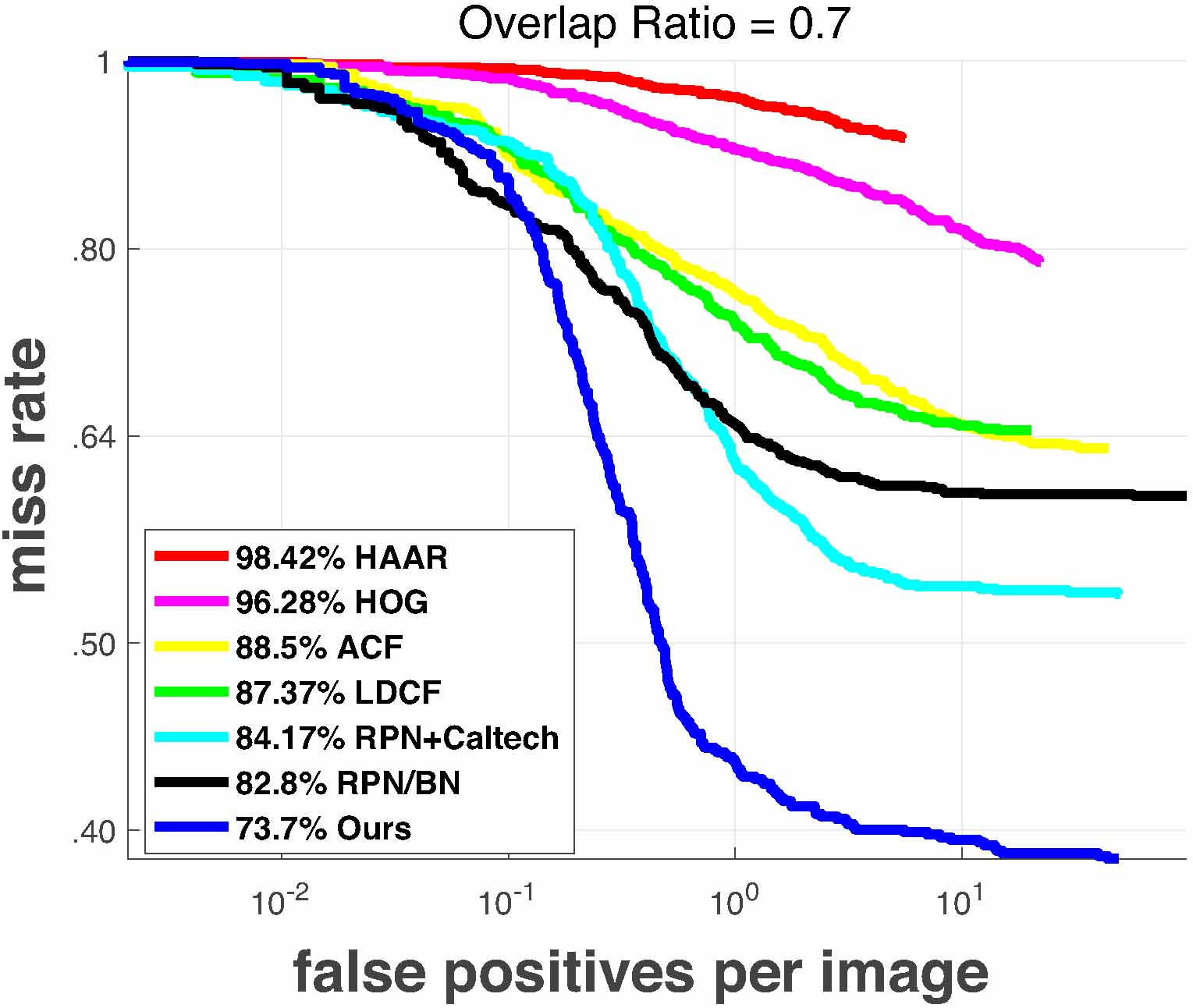}
\end{centering}
}

\end{center}
\vspace{-3ex}
\caption{(a) and (b) are ROC curves for different detectors under different overlap ratio criteria on the Precarious Pedestrian testset. In the legend, we denote the miss rate at $10^{-1}$ false positives per image. {\bf RPN+Caltech} refers to our RPN+ network architecture trained only on Caltech, while {\bf Ours} refers to our detector (RPN+) trained on synthetic, imposter, and real images (Alg~\ref{algo:adapt}). Note all detectors besides {\bf Ours} are trained on the Caltech Dataset.}
\label{fig:precarious}
\end{figure}

\begin{figure*}[t!]
   \subfloat[RPN+Caltech]{
\includegraphics[width=0.5\linewidth]{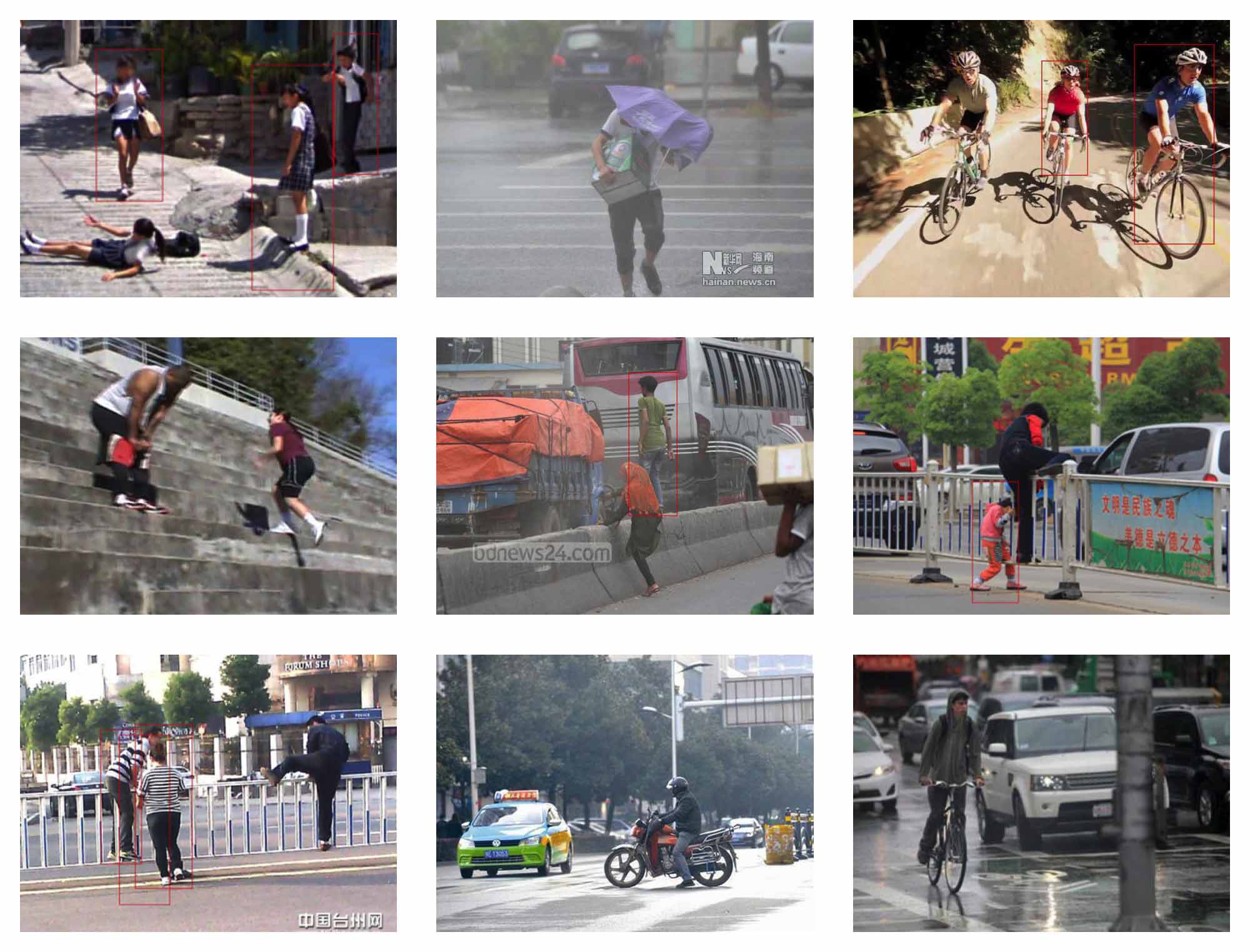}
}\subfloat[Ours]{
\includegraphics[width=0.5\linewidth]{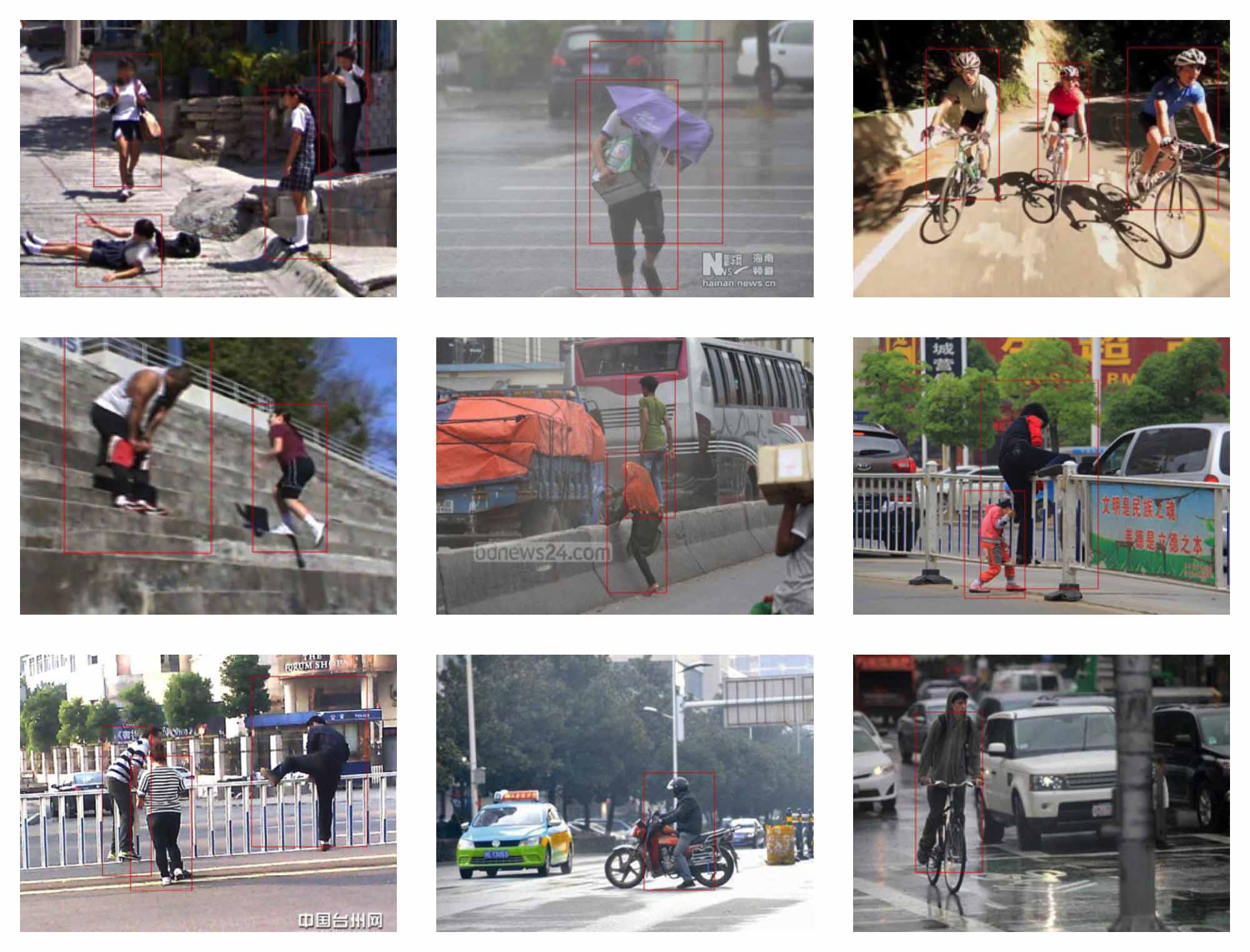}
}
\vspace{-2ex}
   \caption{{\bf Results on Precarious Dataset.} The left shows results of RPN+ trained on Caltech, while the right shows RPN+ trained with adversarial imposters.}
\label{fig:results}
\end{figure*}

\noindent {\bf Precarious Pedestrians:} Results on Precarious Pedestrians are presented in Figure~\ref{fig:precarious}. Our detector significantly outperforms alternative approaches, including the state-of-the-art RPN/BF model. At $10^{-1}$ false positive per image, our miss rate of 42.47\% significantly outperforms all baselines, including the state-of-the-art RPN/BF model (with a miss rate of 54.5\%). Note that all baseline detectors are trained on Caltech. Comparing to baselines is complicated by the fact that both the detection system and training dataset have changed. However, in some sense, our fundamental contribution is method for generating more accurate training datasets through adversarial imposters. To isolate the impact of our underlying detection network RPN+, we also train a variant solely on the Caltech training set (denoted as RPN+Caltech), making it directly comparable to all baselines because they use the same training set. RPN+Caltech performs slightly worse than RPN/BF (with miss-rate of 58.82\%), though it outperforms RPN/BF at higher false positive rates. This suggests that our underlying network is close to state-of-the-art, and moreover validates the significant improvement of training with Adversarial Imposters.  Figure \ref{fig:results} visualizes the results of RPN+, both trained on Caltech and trained with Adversarial Imposters. Qualitatively, we find that Precarious Pedestrians tend to take on more pose variation than typical pedestrians. This requires detection systems that are able to report back a wider range of bounding box scales and aspect ratios.

\begin{figure}[h]
\subfloat[Precarious Dataset]{
\includegraphics[width=0.5\linewidth]{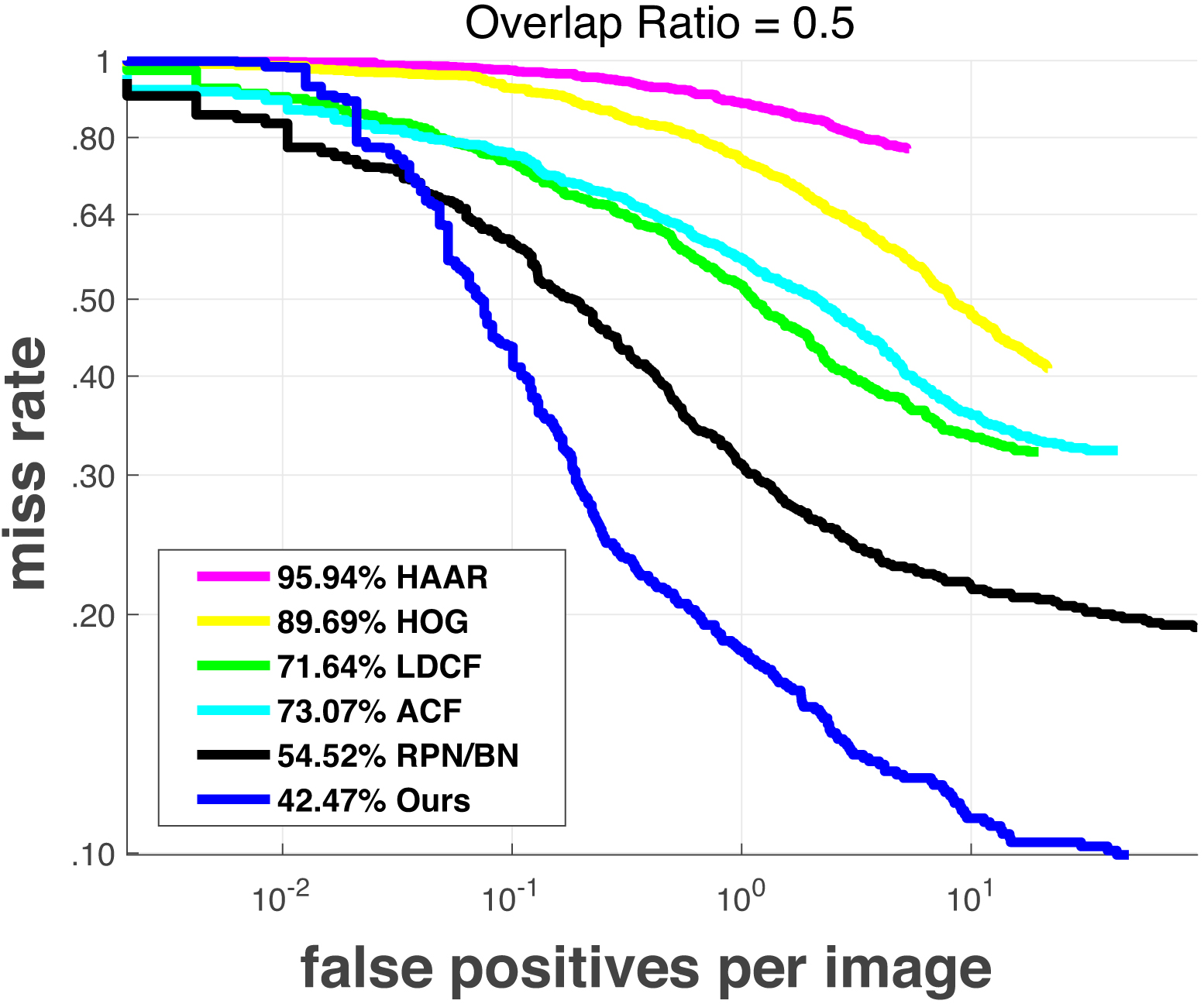}
}
\subfloat[Synthetic Dataset]{
\includegraphics[width=0.5\linewidth]{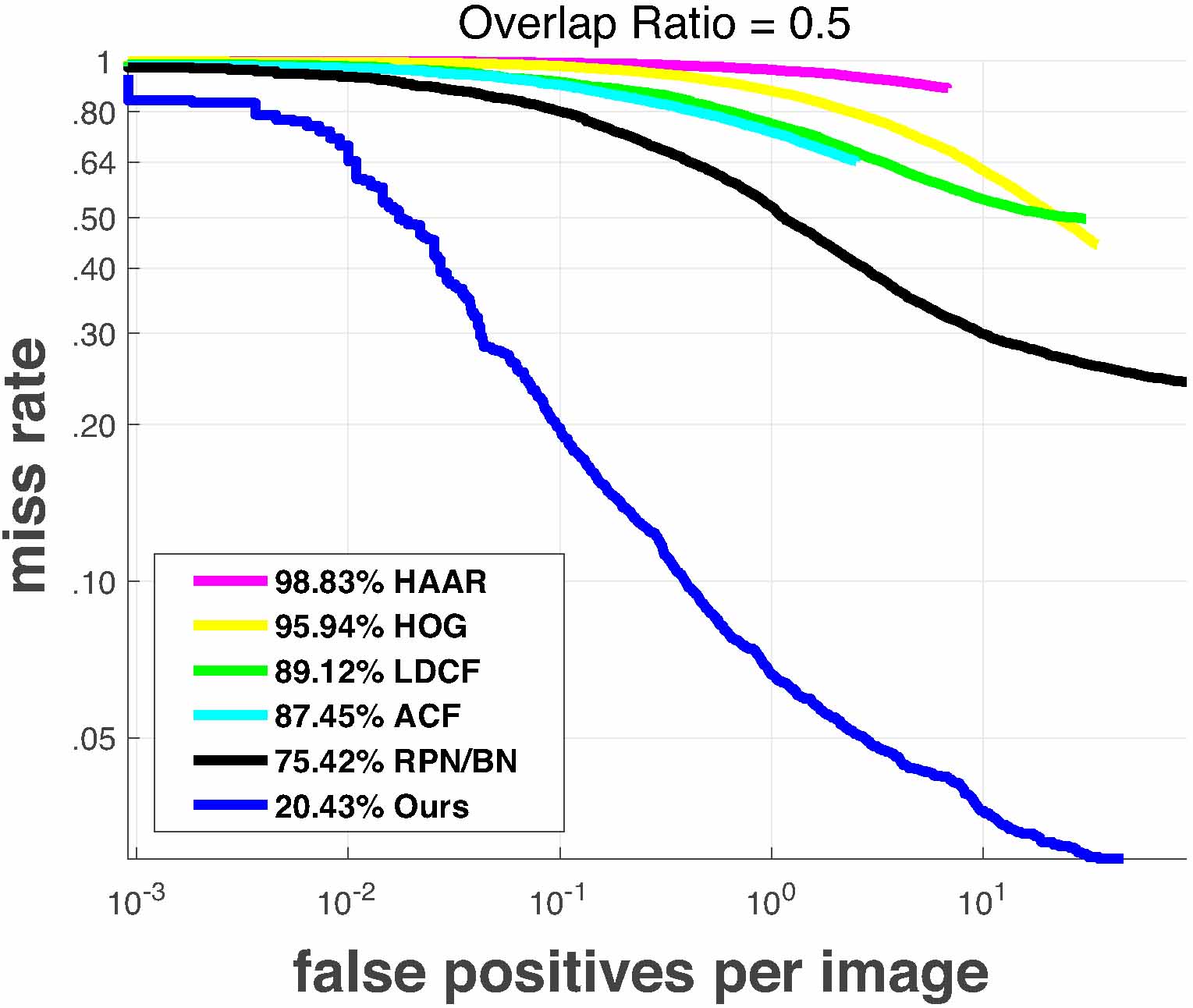}
}
\caption{{\bf Algorithm rankings on real datasets and synthetic dataset.} ROC curves for detectors on Precarious Dataset and Synthetic Dataset. {\bf Ours}(RPN+) is trained with selection model. 
The results suggest that performances of algorithms on real dataset and synthetic dataset have the same rankings.}
\label{fig:compapre}
\end{figure}

\begin{figure}[t!]
\subfloat[Number of imposters]{
\includegraphics[width=0.49\linewidth]{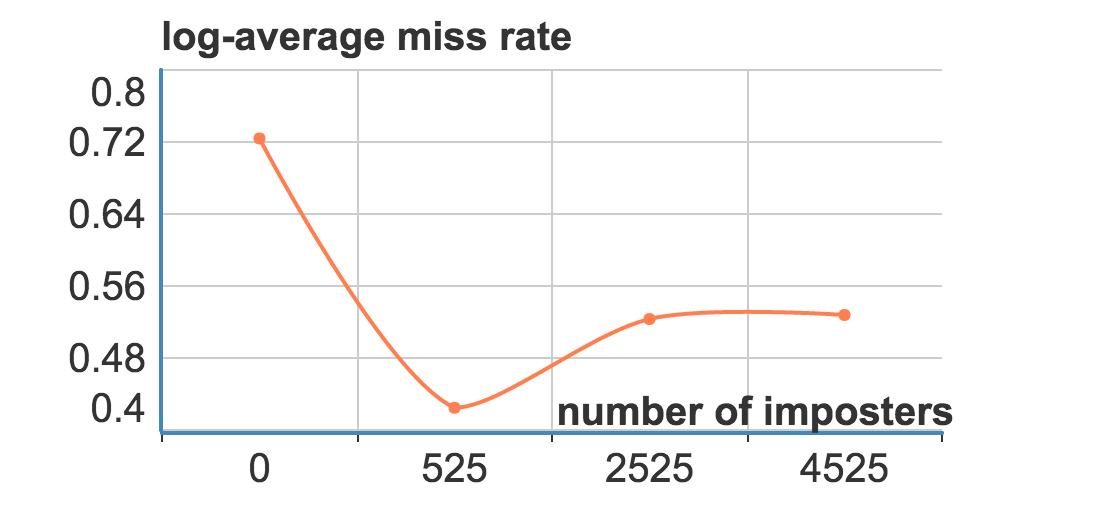}
 
}\subfloat[Number of epochs]{
\includegraphics[width=0.49\linewidth]{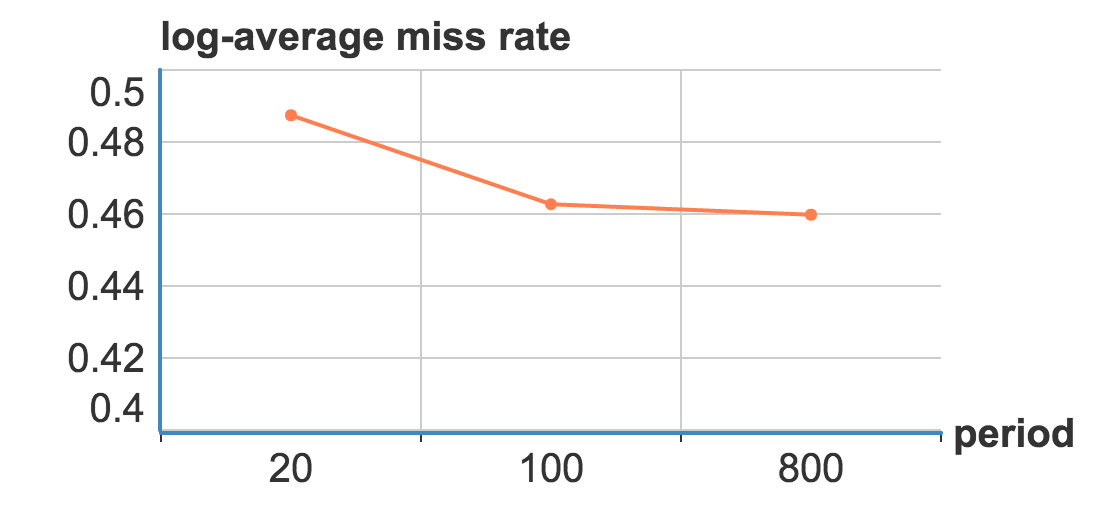}
}
\caption{(a) {\bf Results of different amounts of Imposters.} Optimal performance is obtained with an imposter set that is roughly equal in size to the set of  500 target samples of Precarious training images. (b) {\bf Results of different selectors.} We choose three selectors from different training periods and use them to select imposters separately. As the discriminator gets better, so does the final fine-tuned detector.}
\label{fig:numandimposters}
\end{figure}

\noindent {\bf Adversarial Imposters:} We also explore how detectors perform on a testset of Adversarial Imposters. Note that we can generate an arbitrarily large testset since it is synthetic. 
Figure \ref{fig:compapre} and Figure \ref{fig:numandimposters} show that the performance on both real test data and synthetic test data has the same ranking order. These results suggest that synthetic data may be useful as a testset for evaluating detectors on rare (but important) scenarios that are difficult to observe in real test data. 

\begin{figure}[h]
\begin{center}
\subfloat[Caltech Dataset]{\begin{centering}
\includegraphics[width=0.5\linewidth]{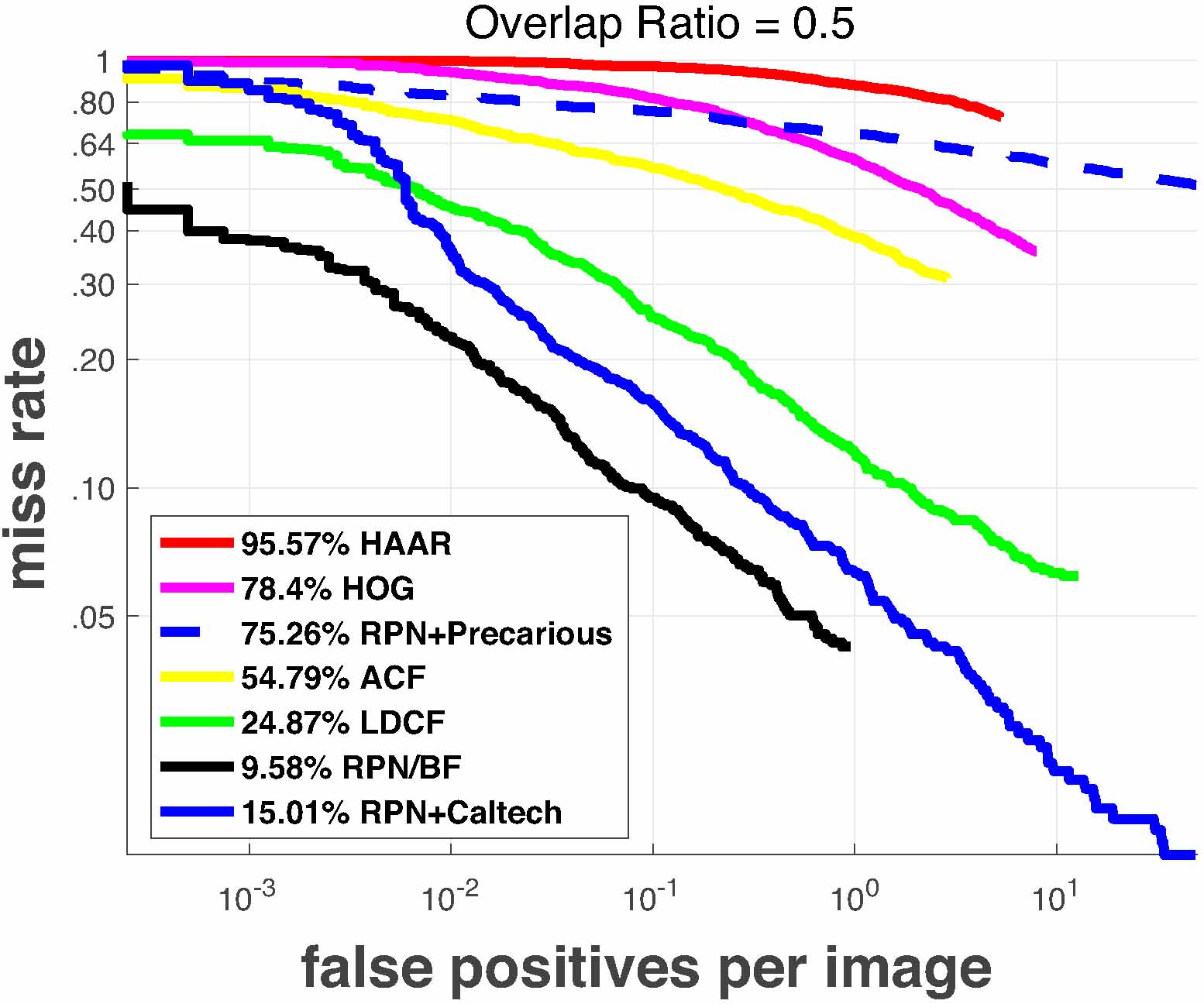}
\end{centering}
}\subfloat[Caltech Dataset]{\begin{centering}
\includegraphics[width=0.5\linewidth]{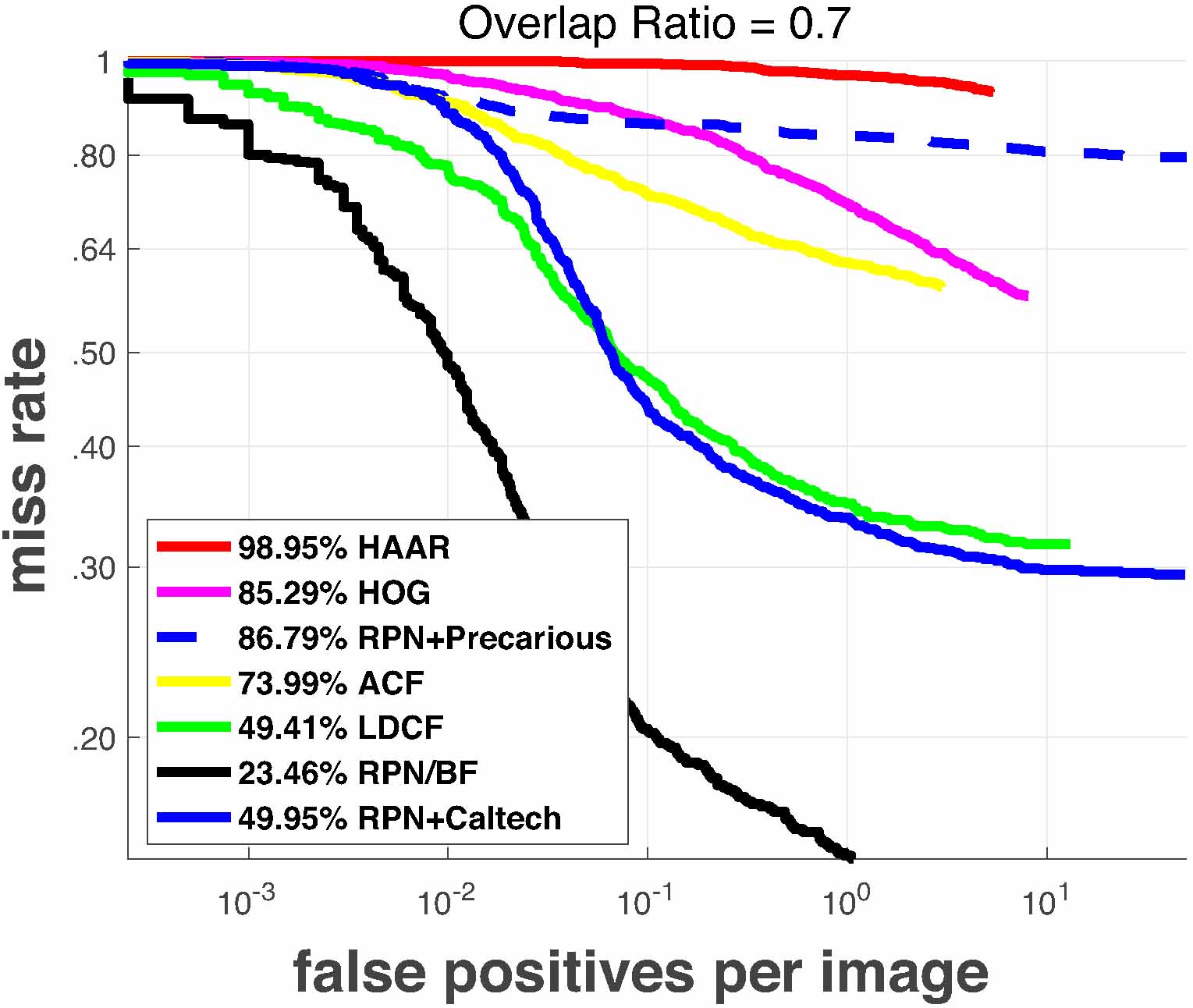}
\end{centering}
}
\end{center}
\vspace{-3ex}
\caption{(a) and (b) are ROC curves for different detectors under different overlap ratio criteria on Caltech Dataset under the default ``reasonable" test protocol.}
\label{fig:caltech}
\end{figure}

\noindent {\bf Caltech:} Finally, for completeness, we also test our RPN+ network on the Caltech Dataset in Figure~\ref{fig:caltech}. Here, all the detectors are trained on Caltech Dataset. For reference, RPN+Caltech model would currently rank 6th out of 68 entries on the Caltech Dataset leaderboard. We also attempted to evaluate our final model (trained with Adversarial Imposters) on Caltech, but saw lackluster performance. We posit that this is due to the different set of scales and aspect ratios in Precarious Pedestrians. We leave further cross-dataset analysis to future work.

\begin{table}[t!]
\begin{center}
\begin{tabular}{|c|c|c|}
\hline
Fine-tuning method & 50\% overlap & 70\% overlap \\
\hline\hline
$S$ & 83.49\% & 95.18\%\\
$T$ & 72.39\% & 93.70\% \\
$S \Rightarrow T$ & 48.45\% & 77.14\% \\
$S \Rightarrow (T \cup I)$ & 45.97\% & 74.94\% \\
$S \Rightarrow (T \cup I) \Rightarrow T$ & \bf{42.47\%} & \bf{73.70}\% \\
\hline
\end{tabular}
\end{center}
\vspace{-3ex}
\caption{Miss rate of different fine-tuning strategies at a false positive rate of $10^{-1}$, where $S$, $T$, and $I$ refer to source datasets (of synthetic images), target dataset (of 
recarious real images), and Imposter dataset.}
\label{table:process}
\end{table}

\subsection{Diagnostics}
In this section, we explore various variants of our approach.  Table \ref{table:process} examines different fine-tuning strategies for adapting detectors from the source domain of synthetic images to the target domain of real Precarious images. Fine-tuning via Imposters performs the best 42.47\%, and noticeably outperforms the commonplace baselines of traditional fine-tuning (by 6\%)  and training on only the target (by 24\%).\\
\indent
Figure \ref{fig:numandimposters} examines the effect of $k$, the size of the imposter set. We find good performance when $k$ is equal to $|T|$, the size of the target set of 
Precarious Pedestrians used for training. In retrospect, this may not be surprising as this produces a balanced distribution of real images and Adversarial Imposters for training. Finally, Figure \ref{fig:numandimposters} also explores the impact of the discriminator. It plots performance as a function of the training epoch used to learn $D({\bf x})$. As we train a better discriminator, the performance of our overall adversarial pipeline gets noticeably better.


\section{Conclusion}
We have explored methods for analyzing ``in-the-tail'' urban scenes, which represent important modes of operations for autonomous vehicles. Motivated by the fact that rare but dangerous scenes are exactly the scenarios on which visual recognition should excel, we first analyze existing datasets and illustrate that they do not contain sufficient rare scenarios (because they naturally focus on common or typical urban scenes). To address this gap, we have collected our own dataset of Precarious Pedestrians, which we will release to spur further research on this important (but under explored) problem. Precarious scenes are challenging because little data is available for both evaluation and training. To address this challenge, we propose the use of synthetic  data generated with a game engine. However, it is challenging to ensure that the synthesized data matches the statistics of real precarious scenarios. Inspired by generative adversarial networks, we introduce the use of a discriminative classifier (trained to discriminate real vs synthetic data) to implicitly specify this distribution. We then use the synthesized data that fooled the discriminator (the ``synthetic imposters'') to both train and evaluate state-of-the-art, robust pedestrian detection systems.~\\

\noindent{\bf Acknowledgements.} This work was supported by NSF Grant 1618903, NSF Grant 1208598, and Google. We thank Yinpeng Dong and Mingsheng Long for helpful discussions.

{\small
\bibliographystyle{ieee}
\bibliography{egbib}

\begin{thebibliography}{10}\itemsep=-1pt

\bibitem{agarwal2006local}
A.~Agarwal and B.~Triggs.
\newblock A local basis representation for estimating human pose from cluttered
  images.
\newblock In {\em Asian Conference on Computer Vision}, pages 50--59. Springer,
  2006.

\bibitem{andriluka14cvpr}
M.~Andriluka, L.~Pishchulin, P.~Gehler, and B.~Schiele.
\newblock 2d human pose estimation: New benchmark and state of the art
  analysis.
\newblock In {\em IEEE Conference on Computer Vision and Pattern Recognition
  (CVPR)}, June 2014.

\bibitem{athitsos2010database}
V.~Athitsos, H.~Wang, and A.~Stefan.
\newblock A database-based framework for gesture recognition.
\newblock {\em Personal and Ubiquitous Computing}, 14(6):511--526, 2010.

\bibitem{broggi2005model}
A.~Broggi, A.~Fascioli, P.~Grisleri, T.~Graf, and M.~Meinecke.
\newblock Model-based validation approaches and matching techniques for
  automotive vision based pedestrian detection.
\newblock In {\em 2005 IEEE Computer Society Conference on Computer Vision and
  Pattern Recognition (CVPR'05)-Workshops}, pages 1--1. IEEE, 2005.

\bibitem{chen2016infogan}
X.~Chen, Y.~Duan, R.~Houthooft, J.~Schulman, I.~Sutskever, and P.~Abbeel.
\newblock Infogan: Interpretable representation learning by information
  maximizing generative adversarial nets.
\newblock {\em arXiv preprint arXiv:1606.03657}, 2016.

\bibitem{dalal2005inria}
N.~Dalal and B.~Triggs.
\newblock Inria person dataset, 2005.

\bibitem{denton2015deep}
E.~L. Denton, S.~Chintala, R.~Fergus, et~al.
\newblock Deep generative image models using a￼ laplacian pyramid of
  adversarial networks.
\newblock In {\em Advances in neural information processing systems}, pages
  1486--1494, 2015.

\bibitem{dollar2014fast}
P.~Doll{\'a}r, R.~Appel, S.~Belongie, and P.~Perona.
\newblock Fast feature pyramids for object detection.
\newblock {\em IEEE Transactions on Pattern Analysis and Machine Intelligence},
  36(8):1532--1545, 2014.

\bibitem{dollar2009pedestrian}
P.~Doll{\'a}r, C.~Wojek, B.~Schiele, and P.~Perona.
\newblock Pedestrian detection: A benchmark.
\newblock In {\em Computer Vision and Pattern Recognition, 2009. CVPR 2009.
  IEEE Conference on}, pages 304--311. IEEE, 2009.

\bibitem{Dollar2012PAMI}
P.~Doll\'ar, C.~Wojek, B.~Schiele, and P.~Perona.
\newblock Pedestrian detection: An evaluation of the state of the art.
\newblock {\em PAMI}, 34, 2012.

\bibitem{eberly20063d}
D.~H. Eberly.
\newblock {\em 3D game engine design: a practical approach to real-time
  computer graphics}.
\newblock CRC Press, 2006.

\bibitem{enzweiler2009monocular}
M.~Enzweiler and D.~M. Gavrila.
\newblock Monocular pedestrian detection: Survey and experiments.
\newblock {\em IEEE transactions on pattern analysis and machine intelligence},
  31(12):2179--2195, 2009.

\bibitem{eth_biwi_00534}
A.~Ess, B.~Leibe, K.~Schindler, , and L.~van Gool.
\newblock A mobile vision system for robust multi-person tracking.
\newblock In {\em IEEE Conference on Computer Vision and Pattern Recognition
  (CVPR'08)}. IEEE Press, June 2008.

\bibitem{fischer2015flownet}
P.~Fischer, A.~Dosovitskiy, E.~Ilg, P.~H{\"a}usser, C.~Haz{\i}rba{\c{s}},
  V.~Golkov, P.~van~der Smagt, D.~Cremers, and T.~Brox.
\newblock Flownet: Learning optical flow with convolutional networks.
\newblock {\em arXiv preprint arXiv:1504.06852}, 2015.

\bibitem{ganin2014unsupervised}
Y.~Ganin and V.~Lempitsky.
\newblock Unsupervised domain adaptation by backpropagation.
\newblock {\em arXiv preprint arXiv:1409.7495}, 2014.

\bibitem{ganin2016domain}
Y.~Ganin, E.~Ustinova, H.~Ajakan, P.~Germain, H.~Larochelle, F.~Laviolette,
  M.~Marchand, and V.~Lempitsky.
\newblock Domain-adversarial training of neural networks.
\newblock {\em Journal of Machine Learning Research}, 17(59):1--35, 2016.

\bibitem{goodfellow2014generative}
I.~Goodfellow, J.~Pouget-Abadie, M.~Mirza, B.~Xu, D.~Warde-Farley, S.~Ozair,
  A.~Courville, and Y.~Bengio.
\newblock Generative adversarial nets.
\newblock In {\em Advances in Neural Information Processing Systems}, pages
  2672--2680, 2014.

\bibitem{grauman2003inferring}
K.~Grauman, G.~Shakhnarovich, and T.~Darrell.
\newblock Inferring 3d structure with a statistical image-based shape model.
\newblock In {\em Computer Vision, 2003. Proceedings. Ninth IEEE International
  Conference on}, pages 641--647. IEEE, 2003.

\bibitem{hattori2015learning}
H.~Hattori, V.~Naresh~Boddeti, K.~M. Kitani, and T.~Kanade.
\newblock Learning scene-specific pedestrian detectors without real data.
\newblock In {\em Proceedings of the IEEE Conference on Computer Vision and
  Pattern Recognition}, pages 3819--3827, 2015.

\bibitem{hejrati2014analysis}
M.~Hejrati and D.~Ramanan.
\newblock Analysis by synthesis: 3d object recognition by object
  reconstruction.
\newblock In {\em 2014 IEEE Conference on Computer Vision and Pattern
  Recognition}, pages 2449--2456. IEEE, 2014.

\bibitem{isola2016image}
P.~Isola, J.-Y. Zhu, T.~Zhou, and A.~A. Efros.
\newblock Image-to-image translation with conditional adversarial networks.
\newblock {\em arXiv preprint arXiv:1611.07004}, 2016.

\bibitem{kingma2013auto}
D.~P. Kingma and M.~Welling.
\newblock Auto-encoding variational bayes.
\newblock {\em arXiv preprint arXiv:1312.6114}, 2013.

\bibitem{lai2014unsupervised}
K.~Lai, L.~Bo, and D.~Fox.
\newblock Unsupervised feature learning for 3d scene labeling.
\newblock In {\em 2014 IEEE International Conference on Robotics and Automation
  (ICRA)}, pages 3050--3057. IEEE, 2014.

\bibitem{lerer2016learning}
A.~Lerer, S.~Gross, and R.~Fergus.
\newblock Learning physical intuition of block towers by example.
\newblock {\em arXiv preprint arXiv:1603.01312}, 2016.

\bibitem{lienhart2002extended}
R.~Lienhart and J.~Maydt.
\newblock An extended set of haar-like features for rapid object detection.
\newblock In {\em Image Processing. 2002. Proceedings. 2002 International
  Conference on}, volume~1, pages I--900. IEEE, 2002.

\bibitem{liu2016coupled}
M.-Y. Liu and O.~Tuzel.
\newblock Coupled generative adversarial networks.
\newblock In {\em Advances in Neural Information Processing Systems}, pages
  469--477, 2016.

\bibitem{long2015learning}
M.~Long, Y.~Cao, J.~Wang, and M.~I. Jordan.
\newblock Learning transferable features with deep adaptation networks.
\newblock In {\em ICML}, pages 97--105, 2015.

\bibitem{loper2014opendr}
M.~M. Loper and M.~J. Black.
\newblock Opendr: An approximate differentiable renderer.
\newblock In {\em European Conference on Computer Vision}, pages 154--169.
  Springer, 2014.

\bibitem{marin2010learning}
J.~Marin, D.~V{\'a}zquez, D.~Ger{\'o}nimo, and A.~M. L{\'o}pez.
\newblock Learning appearance in virtual scenarios for pedestrian detection.
\newblock In {\em Computer Vision and Pattern Recognition (CVPR), 2010 IEEE
  Conference on}, pages 137--144. IEEE, 2010.

\bibitem{mayer2015large}
N.~Mayer, E.~Ilg, P.~H{\"a}usser, P.~Fischer, D.~Cremers, A.~Dosovitskiy, and
  T.~Brox.
\newblock A large dataset to train convolutional networks for disparity,
  optical flow, and scene flow estimation.
\newblock {\em arXiv preprint arXiv:1512.02134}, 2015.

\bibitem{mirza2014conditional}
M.~Mirza and S.~Osindero.
\newblock Conditional generative adversarial nets.
\newblock {\em arXiv preprint arXiv:1411.1784}, 2014.

\bibitem{movshovitz20143d}
Y.~Movshovitz-Attias, Y.~Sheikh, V.~N. Boddeti, and Z.~Wei.
\newblock 3d pose-by-detection of vehicles via discriminatively reduced
  ensembles of correlation filters.
\newblock In {\em BMVC}, 2014.

\bibitem{nam2014local}
W.~Nam, P.~Doll{\'a}r, and J.~H. Han.
\newblock Local decorrelation for improved pedestrian detection.
\newblock In {\em Advances in Neural Information Processing Systems}, pages
  424--432, 2014.

\bibitem{pepik2012teaching}
B.~Pepik, M.~Stark, P.~Gehler, and B.~Schiele.
\newblock Teaching 3d geometry to deformable part models.
\newblock In {\em Computer Vision and Pattern Recognition (CVPR), 2012 IEEE
  Conference on}, pages 3362--3369. IEEE, 2012.

\bibitem{pishchulin2011learning}
L.~Pishchulin, A.~Jain, C.~Wojek, M.~Andriluka, T.~Thorm{\"a}hlen, and
  B.~Schiele.
\newblock Learning people detection models from few training samples.
\newblock In {\em Computer Vision and Pattern Recognition (CVPR), 2011 IEEE
  Conference on}, pages 1473--1480. IEEE, 2011.

\bibitem{potamias2008nearest}
M.~Potamias and V.~Athitsos.
\newblock Nearest neighbor search methods for handshape recognition.
\newblock In {\em Proceedings of the 1st international conference on PErvasive
  Technologies Related to Assistive Environments}, page~30. ACM, 2008.

\bibitem{radford2015unsupervised}
A.~Radford, L.~Metz, and S.~Chintala.
\newblock Unsupervised representation learning with deep convolutional
  generative adversarial networks.
\newblock {\em arXiv preprint arXiv:1511.06434}, 2015.

\bibitem{ren2015faster}
S.~Ren, K.~He, R.~Girshick, and J.~Sun.
\newblock Faster r-cnn: Towards real-time object detection with region proposal
  networks.
\newblock In {\em Advances in neural information processing systems}, pages
  91--99, 2015.

\bibitem{rezende2014stochastic}
D.~J. Rezende, S.~Mohamed, and D.~Wierstra.
\newblock Stochastic backpropagation and approximate inference in deep
  generative models.
\newblock {\em arXiv preprint arXiv:1401.4082}, 2014.

\bibitem{richter2016playing}
S.~R. Richter, V.~Vineet, S.~Roth, and V.~Koltun.
\newblock Playing for data: Ground truth from computer games.
\newblock {\em arXiv preprint arXiv:1608.02192}, 2016.

\bibitem{romero2010hands}
J.~Romero, H.~Kjellstr{\"o}m, and D.~Kragic.
\newblock Hands in action: real-time 3d reconstruction of hands in interaction
  with objects.
\newblock In {\em Robotics and Automation (ICRA), 2010 IEEE International
  Conference on}, pages 458--463. IEEE, 2010.

\bibitem{ros2016synthia}
G.~Ros, L.~Sellart, J.~Materzynska, D.~Vazquez, and A.~M. Lopez.
\newblock The synthia dataset: A large collection of synthetic images for
  semantic segmentation of urban scenes.
\newblock In {\em Proceedings of the IEEE Conference on Computer Vision and
  Pattern Recognition}, pages 3234--3243, 2016.

\bibitem{russakovsky2015imagenet}
O.~Russakovsky, J.~Deng, H.~Su, J.~Krause, S.~Satheesh, S.~Ma, Z.~Huang,
  A.~Karpathy, A.~Khosla, M.~Bernstein, et~al.
\newblock Imagenet large scale visual recognition challenge.
\newblock {\em International Journal of Computer Vision}, 115(3):211--252,
  2015.

\bibitem{salimans2016improved}
T.~Salimans, I.~Goodfellow, W.~Zaremba, V.~Cheung, A.~Radford, and X.~Chen.
\newblock Improved techniques for training gans.
\newblock {\em arXiv preprint arXiv:1606.03498}, 2016.

\bibitem{satkin2012data}
S.~Satkin, J.~Lin, and M.~Hebert.
\newblock Data-driven scene understanding from 3d models.
\newblock 2012.

\bibitem{sun2014virtual}
B.~Sun and K.~Saenko.
\newblock From virtual to reality: Fast adaptation of virtual object detectors
  to real domains.
\newblock In {\em BMVC}, volume~1, page~3, 2014.

\bibitem{viola2001rapid}
P.~Viola and M.~Jones.
\newblock Rapid object detection using a boosted cascade of simple features.
\newblock In {\em Computer Vision and Pattern Recognition, 2001. CVPR 2001.
  Proceedings of the 2001 IEEE Computer Society Conference on}, volume~1, pages
  I--511. IEEE, 2001.

\bibitem{wojek2009multi}
C.~Wojek, S.~Walk, and B.~Schiele.
\newblock Multi-cue onboard pedestrian detection.
\newblock In {\em Computer Vision and Pattern Recognition, 2009. CVPR 2009.
  IEEE Conference on}, pages 794--801. IEEE, 2009.

\bibitem{zhang2016faster}
L.~Zhang, L.~Lin, X.~Liang, and K.~He.
\newblock Is faster r-cnn doing well for pedestrian detection?
\newblock In {\em European Conference on Computer Vision}, pages 443--457.
  Springer, 2016.

\bibitem{zhu2006fast}
Q.~Zhu, M.-C. Yeh, K.-T. Cheng, and S.~Avidan.
\newblock Fast human detection using a cascade of histograms of oriented
  gradients.
\newblock In {\em 2006 IEEE Computer Society Conference on Computer Vision and
  Pattern Recognition (CVPR'06)}, volume~2, pages 1491--1498. IEEE, 2006.

\end{thebibliography}
}

\end{document}